%% file: preprint.tex
\documentclass[journal,twoside,web]{ieeecolor2}
\usepackage{generic_empty}
\usepackage{cite}
\usepackage{amsmath,amssymb,amsfonts}
\usepackage{algorithmic}
\usepackage{graphicx}

\let\labelindent\relax
\usepackage[inline]{enumitem}
\usepackage{bigfoot}
\usepackage{dirtytalk}
\usepackage{pdfpages}

\usepackage{euscript}
\usepackage{MnSymbol}

\usepackage{adjustbox} 
\usepackage[caption=false]{subfig}
\usepackage{float}
\usepackage{dblfloatfix}

\usepackage{booktabs}
\usepackage{makecell}
\usepackage{multirow}
\usepackage{array}

\usepackage[table,xcdraw]{xcolor}
\usepackage{pgf}
\usepackage{color}
\usepackage{tikz}
\usetikzlibrary{shapes}

\makeatletter
\let\NAT@parse\undefined  
\makeatother
\usepackage[breaklinks]{hyperref} 

\makeatletter
\newcommand*{\@rowstyle}{}
\newcommand*{\rowstyle}[1]{
    \gdef\@rowstyle{#1}%
    \leavevmode\@rowstyle
    \ignorespaces
}
\newcolumntype{=}{
    >{\gdef\@rowstyle{}\ignorespaces}%
}
\newcolumntype{+}{
    >{\leavevmode\@rowstyle\ignorespaces}%
}
\makeatother

\def\BibTeX{{\rm B\kern-.05em{\sc i\kern-.025em b}\kern-.08em
    T\kern-.1667em\lower.7ex\hbox{E}\kern-.125emX}}
\pubid{ }

\begin{document}
\title{TUNeS: A Temporal U-Net with Self-Attention for Video-based Surgical Phase Recognition}
\author{Isabel Funke, Dominik Rivoir, Stefanie Krell, and Stefanie Speidel, \IEEEmembership{Member, IEEE}
\thanks{Manuscript received 8 May 2024; revised 4 November 2024 and 15~January 2025; accepted 19 January 2025.
Partially funded by the German Research Foundation (DFG, Deutsche Forschungsgemeinschaft) as part of Germany’s Excellence Strategy – EXC 2050/1 – Project ID 390696704 – Cluster of Excellence ``Centre for Tactile Internet with Human-in-the-Loop'' (CeTI) of Technische Universität Dresden.
The authors acknowledge the financial support by the Federal Ministry of Education and Research of Germany in the programme of ``Souverän. Digital. Vernetzt.''. Joint project 6G-life, project identification number: 16KISK001K.
\emph{(Corresponding author: Isabel Funke.)}
}
\thanks{I. Funke, D. Rivoir, S. Krell, and S. Speidel are with the Department of Translational Surgical Oncology, National Center for Tumor Diseases (NCT/UCC) Dresden, Dresden, Germany; German Cancer Research Center (DKFZ), Heidelberg, Germany; Faculty of Medicine and University Hospital Carl Gustav Carus, TUD Dresden University of Technology, Dresden, Germany; Helmholtz-Zentrum Dresden-Rossendorf (HZDR), Dresden, Germany. }
\thanks{I. Funke, D. Rivoir, and S. Speidel are with the Centre for Tactile Internet with Human-in-the-Loop (CeTI), TUD Dresden University of Technology, Dresden, Germany.}
\thanks{S. Krell and S. Speidel are with the BMBF Research Hub 6G-Life, TUD Dresden University of Technology, Dresden, Germany.
}
}

\maketitle

\bstctlcite{IEEEexample:BSTcontrol}

\begin{abstract}
\emph{Objective:} To enable context-aware computer assistance in the operating room of the future, cognitive systems need to understand automatically which surgical phase is being performed by the medical team.
The primary source of information for surgical phase recognition is typically video, which presents two challenges: extracting meaningful features from the video stream and effectively modeling temporal information in the sequence of visual features.
\emph{Methods:} 
For temporal modeling, attention mechanisms have gained popularity due to their ability to capture long-range dependencies.
In this paper, we explore design choices for attention in existing temporal models for surgical phase recognition and propose a novel approach that uses attention more effectively and does not require hand-crafted constraints:
TUNeS, an efficient and simple temporal model that incorporates self-attention at the core of a convolutional U-Net structure.
In addition, we propose to train the feature extractor, a standard CNN, together with an LSTM on preferably long video segments, i.e., with long temporal context.
\emph{Results:}
In our experiments, almost all temporal models performed better on top of feature extractors that were trained with longer temporal context.
On these contextualized features, TUNeS achieves state-of-the-art results on the Cholec80 dataset.
\emph{Conclusion:} This study offers new insights on how to use attention mechanisms to build accurate and efficient temporal models for surgical phase recognition.
\emph{Significance:}
Implementing automatic surgical phase recognition is essential to automate the analysis and optimization of surgical workflows and to enable context-aware computer assistance during surgery, thus ultimately improving patient care.
\end{abstract}


\input{figure_definitions}

\section{Introduction}
\label{sec:introduction}

\begin{figure*}
	\centerline{%
    \begin{adjustbox}{clip,trim=0.15cm 0.1cm 0.06cm 0.1cm,width=0.8\textwidth,keepaspectratio}
        \includegraphics{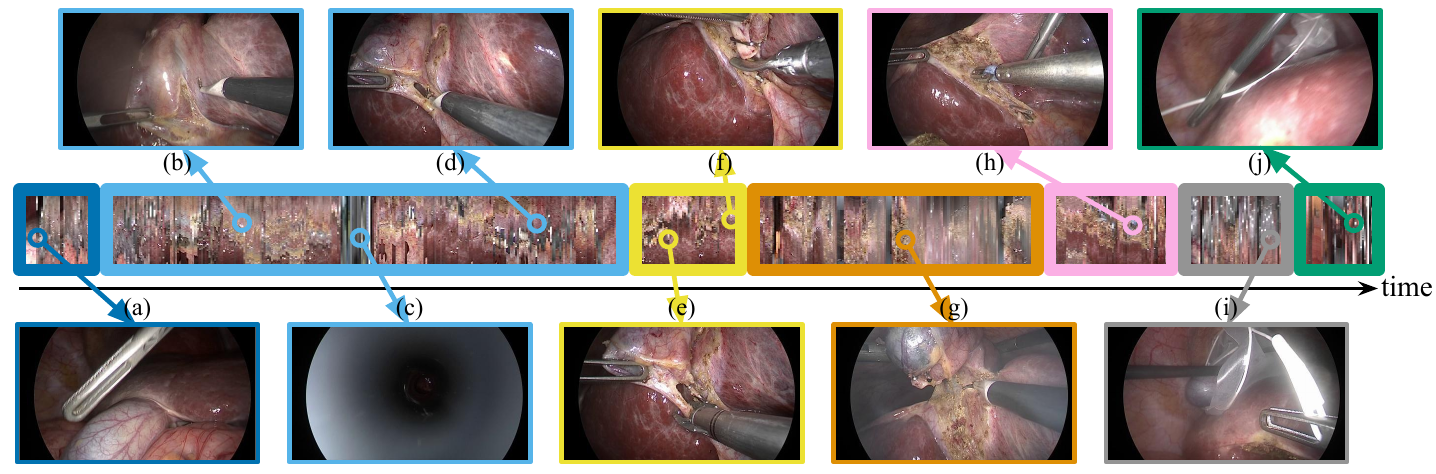}%
    \end{adjustbox}
    }%
    \caption{%
        Visualization of a cholecystectomy video and its segmentation into seven surgical phases \cite{twinanda2016endonet}: \protect\phaseA\,Preparation, \protect\phaseB\,Calot triangle dissection, \protect\phaseC\,Clipping and cutting, \protect\phaseD\,Gallbladder dissection, \protect\phaseE\,Gallbladder packaging, \protect\phaseF\,Cleaning and coagulation, and \protect\phaseG\,Gallbladder retraction. 
        The video is depicted as videogram \cite{davis1995media}, where each video frame is represented by its center strip. (a) -- (j): Representative frames, shown in full spatial resolution.         
    }%
    \label{fig:problem}
\end{figure*}

\IEEEPARstart{V}{ideo-based}
surgical phase recognition refers to automatically analyzing the video stream that is recorded in the operating room (OR) to recognize which surgical phase is being performed by the medical team. This way, the surgical procedure is segmented into consecutive phases, see Fig.~\ref{fig:problem}. 

Here, \emph{surgical phases} correspond to the major events that occur during surgery \cite{lalys2014surgical}. 
For example, a laparoscopic cholecystectomy can be divided into the phases 
\begin{enumerate*}[label=(\arabic*)]
    \item Preparation, 
    \item Calot triangle dissection, where the cystic duct and artery are exposed,
    \item Clipping and cutting, where the cystic duct and artery are transected, 
    \item Gallbladder dissection, where the gallbladder is removed from the liver, 
    \item Gallbladder packaging, where the gallbladder is put into a specimen bag, 
    \item Cleaning and coagulation, where bleedings are coagulated and blood and bile secretion is aspirated, and
    \item Gallbladder retraction, where the specimen bag is removed \cite{twinanda2016endonet}.
\end{enumerate*}

Defining and automatically recognizing surgical phases is applicable to any procedure that follows a standardized workflow. 
Here, the overall number of phases depends on the procedure.
Depending on the procedure, phases could also be executed in different order, be interrupted and resumed, or even be missing in different instances of the same procedure.

\subsection{Clinical motivation}

Recognizing the current phase is a prerequisite for understanding the current surgical situation and context to provide context-aware computer and robot assistance during surgery. 

Examples of \emph{intra-operative} computer assistance include
\begin{enumerate*}[label=(\roman*)]   
    \item clinical decision support, where relevant information is provided to the surgeon at the right time, such as an adaptive visualization of target or at-risk structures \cite{garrow2021machine},
    \item monitoring the surgical workflow so that errors and deviations from the normal operative course are detected early and automatically \cite{garrow2021machine},
    \item context-aware control of surgical robots, e.g., for robotic camera control or robotic scrub nurses \cite{padoy2012statistical}, and
    \item real-time OR scheduling to increase hospital efficiency, such as automatically calling the next patient \cite{padoy2012statistical}.
\end{enumerate*}

\begin{figure}
	\centerline{%
    \begin{adjustbox}{clip,trim=0.15cm 0.07cm 0.12cm 0.15cm,width=0.85\columnwidth,keepaspectratio}
        \includegraphics{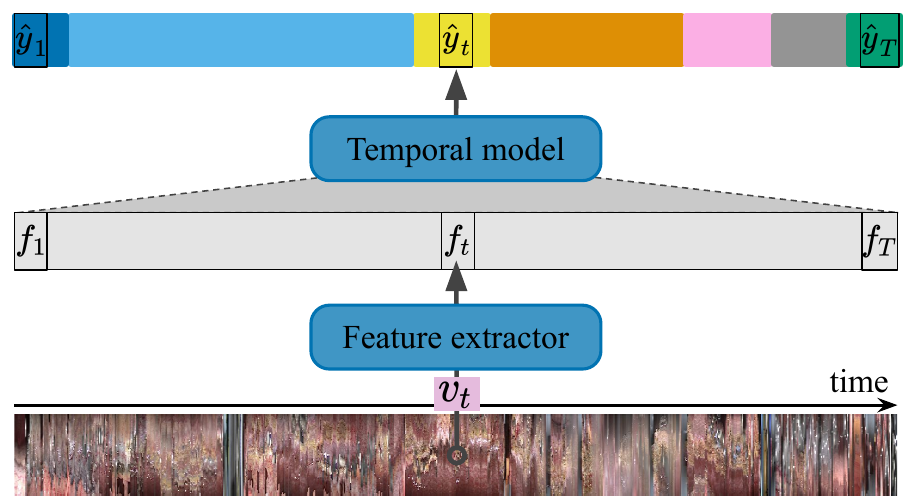}%
    \end{adjustbox}
    }%
    \caption{%
        General framework for video-based surgical phase recognition.
        Whereas the feature extractor computes descriptive features on individual frames, the temporal model considers the relationships with frames that precede (and optionally follow) the current one. 
        (At time~$t$, $v_t$ is the video frame, $f_t$ the feature vector, and $\hat{y}_t$ the phase estimate.)
    }%
    \label{fig:framework}
\end{figure}

In addition, phase recognition forms the foundation for the \emph{post-operative} analysis of surgery recordings. Here, surgical videos are parsed automatically for diverse purposes, including
\begin{enumerate*}[label=(\roman*)]  
    \item documenting the course of the surgery by generating a partially filled surgical report \cite{padoy2012statistical},
    \item locating important steps in the video for documenting, for example, the critical view of safety when transecting risk structures during cholecystectomy \cite{mascagni2021computer},  
    \item creating educational material for surgical trainees \cite{twinanda2016endonet}, and
    \item using statistics to model, quantify, and compare surgical workflows with the goal to optimize the use of physical and human resources and standardize processes \cite{lalys2014surgical}.
\end{enumerate*}

\subsection{Formal problem}
Given a video~$v$ of length~$T$, the task is to estimate which phase~$p$ is being executed at any time~$t$.
Here, $p$ is one of $C$~surgical phases. 
We denote the true phase that is happening at time~$t$ as $y_t$, $1 \leq y_t \leq C$, and the sequence of all ground truth labels as $\left(y_t\right)_{1 \leq t \leq T}$ or $y$. The estimate of the phase at time~$t$ is $\hat{y}_t \in \mathbb{R}^C$, where $\hat{y}_{t,p}$ is the score for the event that $y_t$ equals phase~$p$. 
The video itself is a sequence of video frames~$v_t$. 

For intra-operative applications, phase recognition needs to be performed \emph{online}. Here, only video frames $v_{t'}$ up until the current time~$t$, i.e., $t' \leq t$, can be analyzed to estimate the current phase. For post-operative applications, however, \emph{offline} recognition is feasible. Here, information from all frames in the video can be processed to estimate the phase at any time.

\subsection{Technical approach}

Current methods for surgical phase recognition follow a data-driven approach, meaning that they learn from annotated example videos.
In a naive approach, a standard image classifier, e.g., a \emph{Convolutional Neural Network (CNN)} or a \emph{Vision Transformer (ViT)} \cite{dosovitskiy2020image}, could be trained to recognize the phase in each video frame individually. 
However, classifying single frames is difficult due to blur and occlusions caused by smoke, blood, and camera motion. In addition, there is high variability within a phase whereas frames from different phases can be visually similar \cite{twinanda2016endonet,jin2017sv}. Fig.~\ref{fig:problem} shows some examples. 

Considering the temporal context of each frame can help to resolve ambiguities. Thus, it is common to use the image classifier as \emph{visual feature extractor} and combine it with a \emph{temporal model} that analyzes the full sequence of feature vectors to add temporal context and consider temporal dependencies (Fig.~\ref{fig:framework}).
For handling long surgical videos, the temporal model should be able to capture long-range temporal relationships.
Typical choices are \emph{Temporal Convolutional Networks (TCNs)} \cite{lea2017temporal,farha2019ms} and \emph{Recurrent Neural Networks (RNNs)}, which consist of \emph{Long Short-Term Memory (LSTM)} cells \cite{hochreiter1997long}  or \emph{Gated Recurrent Units (GRUs)} \cite{cho2014learning}. 
Recent temporal models also integrate attention mechanisms \cite{bahdanau2014neural,vaswani2017attention}.

\subsubsection{Visual feature extraction}
The question of how to learn suitable features for phase recognition has not been studied extensively so far. Most methods simply finetune a CNN, typically a \emph{ResNet} \cite{he2016deep}, on the task of phase recognition after pre-training on a large-scale dataset.
Few studies \cite{czempiel2022surgical,he2022empirical} investigated other architectures such as ViTs or spatio-temporal video models.
Others explored training a standard CNN either on additional tasks, e.g., tool recognition \cite{twinanda2016endonet,czempiel2020tecno}, surgical step recognition \cite{ramesh2021multi}, or surgical scene segmentation \cite{sanchez2022data}, or on  self-supervised objectives \cite{bodenstedt2017unsupervised,funke2018temporal,ramesh2023dissecting}.

On the other hand, some methods \cite{jin2017sv,rivoir2022pitfalls} train the CNN \emph{with temporal context} by combining it with an LSTM network and training both components jointly on truncated video sequences. 
Training with temporal context rewards the CNN to extract features that could prove useful in hindsight and enables the CNN to tolerate ambiguous frames, relying on the LSTM to resolve these.
Yet, these methods are limited to training on very short (few seconds) video sequences \cite{jin2017sv} or to CNNs without \emph{Batch Normalization (BN)} layers \cite{rivoir2022pitfalls}.

In contrast, \emph{we achieve to train a standard CNN with BN layers with extended temporal context (up to 64~frames) for visual feature extraction} by using a particular, yet simple, strategy (see section~\ref{sec:temporal_context}):
We make sure that the training batches are \emph{diverse} by sampling several (at least ten) 
video sequences per batch from throughout the training data.

Recently, Liu \emph{et al.} \cite{liu2023lovit} presented a similar strategy to train a \say{temporally-rich spatial feature extractor}: They train a ViT as feature extractor jointly with a Transformer decoder as temporal model on sequences of 30 frames.

Due to computational constraints, the temporal context, i.e., the length $L$ of the training sequence, is still limited compared to the full video ($L << T$). 
Thus, it is beneficial to use a global temporal model in addition. 
Here, we use the trained and frozen CNN as feature extractor and discard the LSTM.

\subsubsection{Temporal model to capture long-range relationships}

\begin{figure*}
	\centerline{%
    \begin{adjustbox}{clip,trim=0.cm 0.cm 0.cm 0.cm,width=0.96\textwidth,keepaspectratio}
        \includegraphics{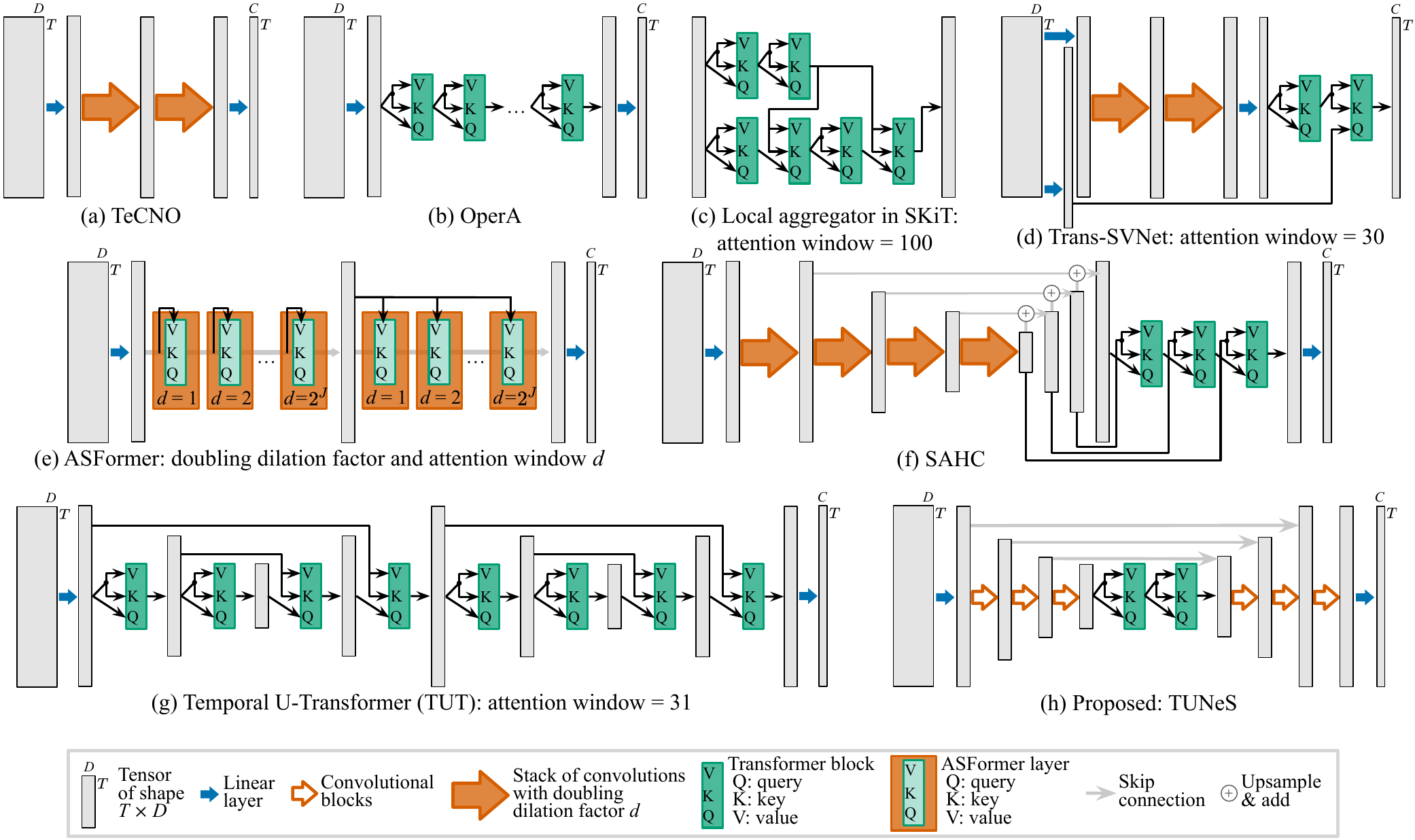}%
    \end{adjustbox}
    }%
    \caption{%
    Temporal models for surgical phase recognition or temporal action segmentation incorporate attention in different ways, see section~\ref{sec:related_work_attention}. ($T$:~length of visual feature sequence, $D$: dimension of visual feature, $C$: number of phases)
    }%
    \label{fig:architectures}
\end{figure*}

\emph{Attention mechanisms} \cite{bahdanau2014neural} were designed to model global relations in sequences. 
By stacking attention and feedforward neural networks, the \emph{Transformer} architecture \cite{vaswani2017attention} achieved break-through results on sequence modeling tasks.
Inspired by this success, researchers incorporated attention in temporal models for phase recognition as well \cite{czempiel2021opera,liu2023skit,gao2021trans,ding2022exploring} (see Fig.~\ref{fig:architectures}, section~\ref{sec:related_work_attention}). 

However, specific challenges apply to the task of phase recognition, similar to the broader field of \emph{temporal action segmentation} \cite{ding2023temporal}: there are inherent constraints on locality and temporal order that need to be modeled, the videos are long and can vary widely in duration, and the available labeled datasets are small. 
Yet, attention mechanisms lack inductive bias and have quadratic time and memory complexity, which makes them computationally expensive on long sequences. 
Thus, using attention effectively and efficiently in temporal models for phase recognition is an area of ongoing research. 

Previous methods chose strategies that constrain the capabilities of attention, namely special losses to control attention weights \cite{czempiel2021opera,du2023dowe} or reverting to \emph{local attention} \cite{liu2023skit,gao2021trans,yi2021asformer,du2023dowe}, which is computed within limited time windows and thus cannot capture global dependencies.
In contrast, \emph{we propose a solution that incorporates global attention efficiently and without imposing hand-crafted constraints}.

Therefore, we build upon the convolutional \emph{U-Net} \cite{ronneberger2015u} encoder-decoder architecture, which was previously proposed for action segmentation \cite{lea2017temporal,singhania2023c2f}
but seemed less popular than multi-stage TCNs (\emph{MS-TCNs})\cite{farha2019ms}. 
A U-Net has the advantage of providing semantically meaningful features at the bottleneck after the encoder, which are also downsampled along the temporal dimension.
We suggest to integrate attention at this position (see Fig.~\ref{fig:architectures}h, section~\ref{sec:tunes}), where attention can be computed efficiently and unconstrained. 
The proposed model, named \emph{TUNeS (\textbf{T}emporal \textbf{U}-\textbf{Ne}t with \textbf{S}elf-Attention)}, is suited for both online and offline phase recognition.

\subsubsection{Evaluation}
We conducted a large number of experiments (section \ref{sec:exp_baselines}) to demonstrate the positive effects of training the feature extractor with long temporal context. Our experiments also show that the temporal model TUNeS compares favorably to many baselines, in terms of both recognition accuracy and computational efficiency. TUNeS, combined with a feature extractor that is trained with long temporal context, achieves state-of-the-art results on Cholec80.
The source code is available at \url{https://gitlab.com/nct_tso_public/tunes}.

\begin{figure}
    \captionsetup[subfloat]{farskip=0cm}
    \centering
        \captionsetup[subfloat]{justification=raggedright,singlelinecheck=false,margin=-1.0cm}
        \subfloat[Transformer block \cite{vaswani2017attention}\label{fig:transformer}]{%
        \begin{adjustbox}{clip,trim=0.15cm 0.cm 0.1cm 0.07cm,width=0.268\columnwidth,keepaspectratio}
        \includegraphics{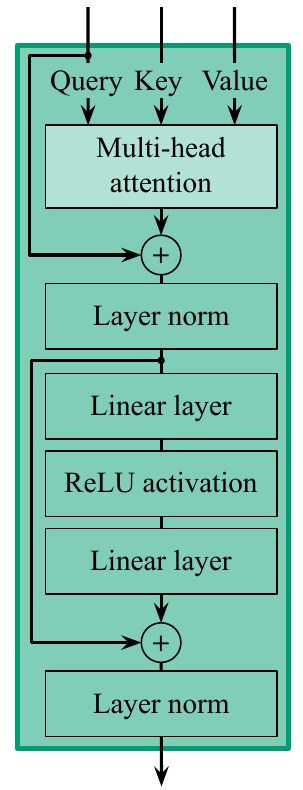}%
        \end{adjustbox}
    }%
    \qquad \qquad
    \subfloat[Layer in MS-TCN \cite{farha2019ms} and ASFormer \cite{yi2021asformer}\label{fig:tcn}]{%
        \begin{adjustbox}{clip,trim=0.cm 0.cm 0.2cm 0.07cm,width=0.285\columnwidth,keepaspectratio}
        \includegraphics{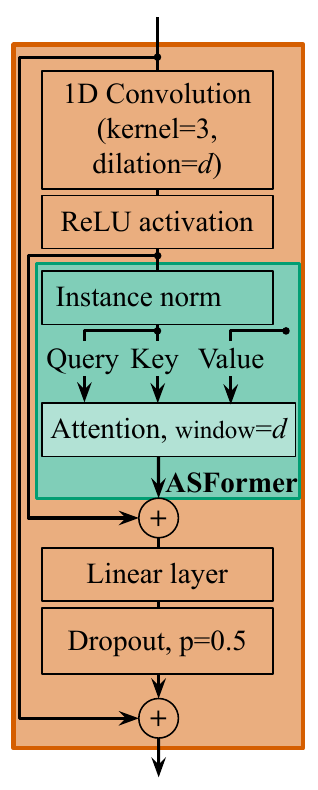}%
        \end{adjustbox}
    }%
    \caption{%
    Blocks of related models. \emph{ASFormer} adds attention between the convolution and the fully connected layer of the \emph{MS-TCN} layer.
    }%
\end{figure}

\section{Attention mechanism}
Let $Q = q_{1:S}$ be a sequence of \emph{queries} and $U = u_{1:T}$ be a sequence of \emph{values}, which is associated with a sequence $K = k_{1:T}$ of \emph{keys}, i.e., $k_t$ is the key of $u_t$. 
In an attention mechanism \cite{bahdanau2014neural}, each element $q_s \in Q$ collects information from any element $u_t \in U$ by computing an attention-weighted sum over the complete sequence~$U$:
\begin{equation}
    \begin{split}
    \mathrm{Attention}(Q, K, U) &= \left(x_s\right)_{1 \leq s \leq S},\text{ where } \\
    x_s &= \sum_t \frac{\exp{\alpha(q_s, k_t)}}{\sum_{t'}\exp{\alpha(q_s, k_{t'})}} \cdot u_t
    \label{eq:attention}
    \end{split}
\end{equation}
The weights in this sum are the attention weights $\alpha(q_s, k_t)$ after Softmax. 
Here, function $\alpha$ is trained to determine how relevant the element with key $k_t$ is given query~$q_s$. 

In practice, $Q$, $K$, and $U$ are passed through trainable linear layers before computing attention, and $\alpha$ simply computes the scaled dot product.
In \emph{multi-head} attention, multiple linear layers are used so that attention is computed on multiple representation subspaces in parallel.
In the case of \emph{self-attention}, $Q$, $K$, and $U$ all refer to the same sequence.  \cite{vaswani2017attention}

Attention weights may be \emph{masked}, i.e., set to $-\infty$, according to specific rules: \emph{Local} attention is only computed within local time windows of size $1 + 2 \omega$, i.e., $\alpha(q_s, k_t) = -\infty$ for all $t$ with $|t - s| > \omega$. \emph{Causal} attention does not allow attending to future sequence elements, thus $\alpha(q_s, k_t) = -\infty$ for all $t$ with $t > s$.

In the Transformer architecture \cite{vaswani2017attention}, attention is embedded in \emph{Transformer blocks} with residual connections and normalization layers (Fig.~\ref{fig:transformer}). Here, the attention mechanism is followed by a pointwise non-linear feedforward network. 

\section{Related work}
\label{sec:related_work}

\subsection{Phase recognition without attention mechanisms}

Early models for video-based phase recognition consisted of a CNN for feature extraction and an LSTM network as temporal model \cite[chapter 6.4]{twinanda2017vision}. Later, Czempiel \emph{et al.} \cite{czempiel2020tecno} proposed \emph{TeCNO} for temporal modeling (Fig.~\ref{fig:architectures}a). TeCNO is a TCN with multiple stages \cite{farha2019ms}, where subsequent stages refine predictions of previous stages. 
Each stage consists of several layers with dilated temporal convolutions (Fig.~\ref{fig:tcn}), where the dilation factor doubles at each layer.
For online recognition, \emph{causal} dilated convolutions \cite{oord2016wavenet} are used, i.e., convolutions are computed over the current and preceding time steps only. 
Yi \emph{et al.} \cite{yi2022not} suggested to train multi-stage models \emph{not end-to-end}, but to apply disturbances to the intermediate predictions during training. 
Recently, Kadkhodamohammadi \emph{et al.} \cite{kadkhodamohammadi2022patg} introduced \emph{PATG}, which uses a SE-ResNet as feature extractor and a graph neural network as temporal model. Zhang \emph{et al.} \cite{zhang2022retrieval} proposed a reinforcement learning approach to retrieve phase transitions for offline recognition. 

Whereas feature extractor and temporal model were trained separately in these studies, 
others proposed to train both models jointly. 
Jin \emph{et al.} \cite{jin2017sv} trained \emph{SV-RCNet}, a ResNet-LSTM, end-to-end on 2\,s long video sequences. Later, they proposed \emph{TMRNet} \cite{jin2021temporal}, which augments a ResNeSt-LSTM with a long-range memory bank \cite{wu2019long} so that feature vectors from the previous 30\,s can be queried in addition. 
Recently, Rivoir \emph{et al.} \cite{rivoir2022pitfalls} showed that CNN-LSTMs should be trained on even longer video sequences, which becomes feasible only if effects caused by BN are taken into account. They trained their best model, a ConvNeXt-LSTM, on sequences of 256\,s.

\subsection{Phase recognition with attention mechanisms}
\label{sec:related_work_attention}

To model long-range dependencies, recent studies proposed to integrate attention mechanisms into temporal models (Fig.~\ref{fig:architectures}).
With attention, an element~$q_s$ in the query sequence can be related to any element~$u_t$ in the value sequence, regardless of their temporal distance~$|t - s|$.

Czempiel \emph{et al.} \cite{czempiel2021opera} presented \emph{OperA}, a stack of Transformer blocks, for online recognition (Fig.~\ref{fig:architectures}b). This approach uses only self-attention to model temporal dependencies and thus may be lacking a notion of locality and temporal order. 
Since the model seemed to struggle to filter relevant information from the feature sequences, the authors proposed to train with a special attention regularization loss. 

Liu \emph{et al.} \cite{liu2023skit} used a 2-layer Transformer encoder-decoder as \emph{local aggregator} in their \emph{SKiT} model for online recognition (Fig.~\ref{fig:architectures}c). Here, the local aggregator considers feature vectors of the previous 100\,s. 
Similar architectures are used for tasks like machine translation or image captioning, where the encoder processes the input and the decoder auto-regressively generates the output, based on the encoded input. In SKiT, however, encoder and decoder process the same information.   
The output of the local aggregator is further refined using a \emph{key information recorder}, which performs max pooling in a low-dimensional space and in a causal manner, i.e., regarding only the current and preceding time steps.

Gao \emph{et al.} \cite{gao2021trans} introduced \emph{Trans-SVNet} for online recognition (Fig.~\ref{fig:architectures}d), which adds two Transformer blocks on top of a frozen TeCNO model. The first block computes self-attention on the TeCNO predictions. In the second block, the initial predictions of the feature extractor attend over the output of the first Transformer block. All attention operations are computed locally, using the feature vectors of the previous 30\,s only. 

Yi \emph{et al.} \cite{yi2021asformer} proposed \emph{ASFormer}, an offline model for action segmentation (Fig.~\ref{fig:architectures}e). This model integrates attention into MS-TCN layers (Fig.~\ref{fig:tcn}), where the convolutions help to encode local inductive bias. Again, attention is computed within local windows. Here, the window size equals the dilation factor~$d$ of the convolution, which doubles at each layer. 
Restricting attention to local neighborhoods of a predefined size may help to handle the complexity of long, noisy feature sequences but hinders the modeling of global dependencies.

Recently, Zhang \emph{et al.} \cite{zhang2022surgical} suggested to add a MS-TCN stage in parallel to the first ASFormer stage and to fuse their outputs. 
Chen \emph{et al.} \cite{chen2022spatio} presented an online two-stage version of ASFormer, where the dilation factors in the second stage halve (instead of double) at each layer. 

Whereas previous models incorporated hierarchical patterns mostly implicitly, e.g. with doubling dilation factors, Ding and Li \cite{ding2022exploring} proposed an explicitly hierarchical model: \emph{SAHC} (Fig.~\ref{fig:architectures}f).
This model uses MS-TCN-like stages to process feature sequences, which are downsampled progressively. Then, the obtained multi-scale feature maps are fused from coarse to fine. 
Transformer blocks with global attention are integrated at the very end, where the final high-resolution sequence queries further information from the lower-resolution sequences. 

Du \emph{et al.} \cite{du2023dowe} presented \emph{Temporal U-Transformer (TUT)}, a hierarchical offline model for action segmentation (Fig.~\ref{fig:architectures}g). TUT consists of Transformer blocks that are equipped with downsampling and upsampling layers to form a U-Net. Here, blocks in the upsampling path obtain information from the downsampling path by attending over the respective feature maps. Multiple U-Nets are stacked to create a multi-stage architecture.  
Local attention is used throughout the model. In addition, a boundary-awareness loss is employed to control the distribution of attention weights around action boundaries.

\section{Methods}
\label{sec:methods}

\subsection{Feature extractor with temporal context}
\label{sec:temporal_context}

\begin{figure}
    \captionsetup[subfloat]{farskip=0cm}
    \centering
    \subfloat[\label{fig:training_no_context}]{%
        \begin{adjustbox}{clip,trim=0.2cm 0.08cm 0.17cm 0.28cm,width=0.18\columnwidth,keepaspectratio}
        \includegraphics{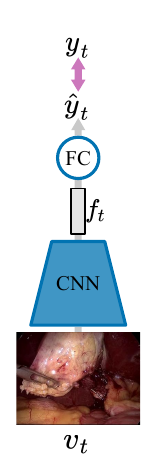}%
        \end{adjustbox}
    }%
    \qquad \qquad
    \subfloat[\label{fig:training_with_context}]{%
        \begin{adjustbox}{clip,trim=0.28cm 0.08cm 0.27cm 0.28cm,width=0.605\columnwidth,keepaspectratio}
        \includegraphics{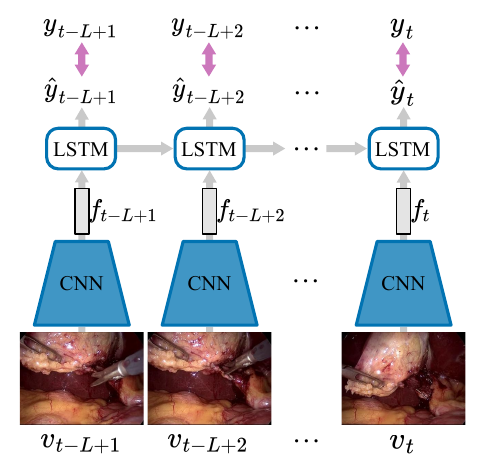}%
        \end{adjustbox}
    }%
    \caption{%
    Feature extractor training: (a) Training the CNN on individual video frames. The fully connected (FC) layer acts as linear classifier on top of the extracted feature vector $f_t$. (b) Training the CNN with temporal context, using an LSTM cell to model temporal relationships.
    }%
    \label{fig:feature_extractor}
\end{figure}

We use a standard CNN as feature extractor: ResNet-50 \cite{he2016deep}, pre-trained on ImageNet. 
However, we train the CNN not on individual video frames (Fig.~\ref{fig:training_no_context}) but with \emph{temporal context}. 
For this purpose, we add an LSTM cell with 512 hidden units, which processes the sequence of computed CNN features. The CNN-LSTM is trained end-to-end on sequences of $L$ consecutive frames (Fig.~\ref{fig:training_with_context}). 
Here, we minimize the cross-entropy loss, computed on all $L$~frames in the sequence. 

To train on a balanced set of video sequences, we sample five sequences from each phase and one sequence around each phase transition from each video during one training epoch.
To avoid BN-related problems, video sequences are processed in appropriately large batches of $B \times L$ frames, $B \geq 10$.

After training, only the CNN is used to extract features: For each frame~$v_t$, we obtain the feature vector~$f_t \in \mathbb{R}^{D}$, $D = 2048$, after the ResNet's global average pooling layer.
To model global dependencies and for offline recognition, a temporal model is trained on the extracted feature sequences $\left(f_t\right)_{1 \leq t \leq T}$ in a second step.

\begin{figure*}
	\centerline{%
    \begin{adjustbox}{clip,trim=0.cm 0.cm 0.06cm 0.cm,width=0.93\textwidth,keepaspectratio}
        \includegraphics{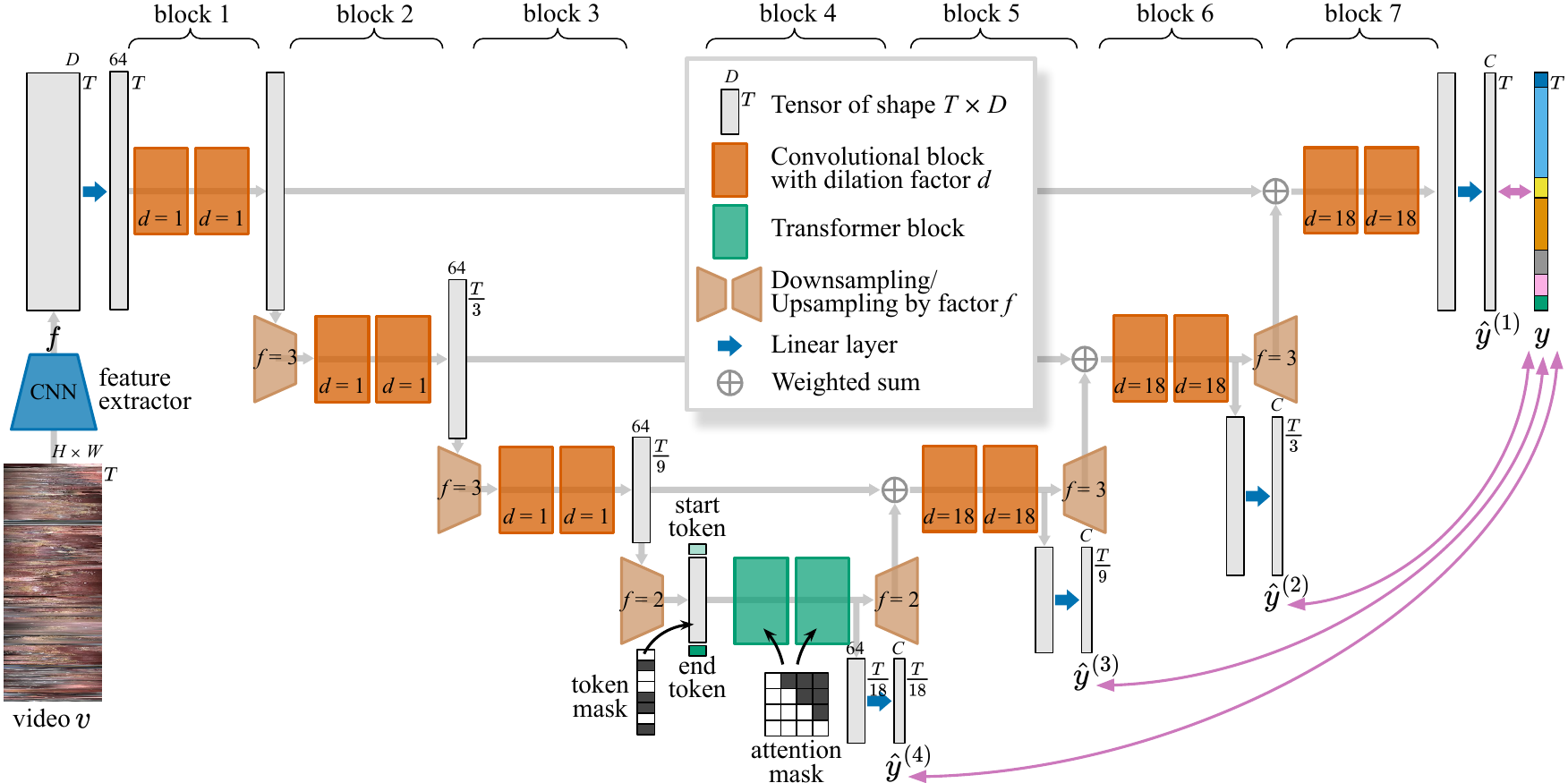}%
    \end{adjustbox}
    }%
    \caption{%
    TUNeS is a U-Net of temporal convolutions, which consists of seven \emph{TUNeS blocks}. To model long-range dependencies, self-attention is used at the bottleneck (\emph{block~4}).    
    TUNeS processes the sequence of visual features $f$, which is obtained by applying the feature extractor (section \ref{sec:temporal_context}) to every frame~$v_t$ (with dimensions $H \times W$) in the video. 
    Multi-scale phase predictions~$\hat{y}^{(i)}$ are extracted from the U-Net decoder to supervise the model during training.
    ($T$: length of video, $D$: dimension of visual feature, $C$: number of phases)
    }%
  \label{fig:overview}
\end{figure*}

\begingroup
\setlength{\tabcolsep}{-2pt}
\begin{figure}
    \centering
    \begin{tabular}{cc}
        \adjustbox{valign=b}{
            \captionsetup[subfloat]{farskip=0cm,justification=raggedright,singlelinecheck=false,margin=1.13cm}
            \subfloat[\label{fig:transformer-block}]{%
                \begin{adjustbox}{clip,trim=0.1cm 0.0cm 0.08cm 0.0cm,width=0.375\columnwidth,keepaspectratio}
                    \includegraphics{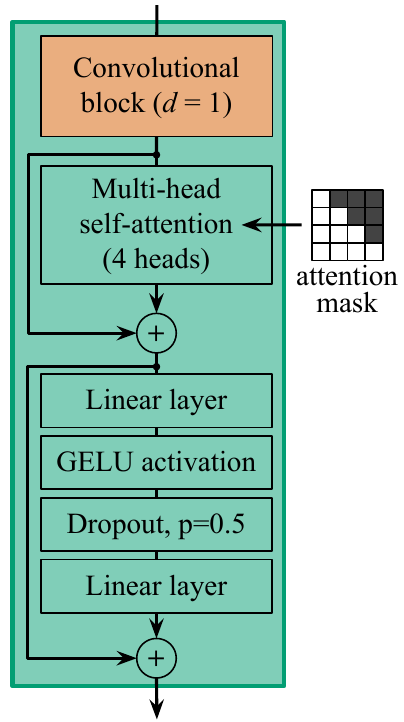}%
                \end{adjustbox}
            }
        }
        &      
        \adjustbox{valign=b}{
            \begin{tabular}{@{}c@{}}
                \captionsetup[subfloat]{farskip=0cm}
                \subfloat[Regular vs. causal operations \label{fig:causal-ops}]{%
                    \begin{adjustbox}{clip,trim=0.cm 0.0cm 0.cm 0.0cm,width=0.605\columnwidth,keepaspectratio}
                        \includegraphics{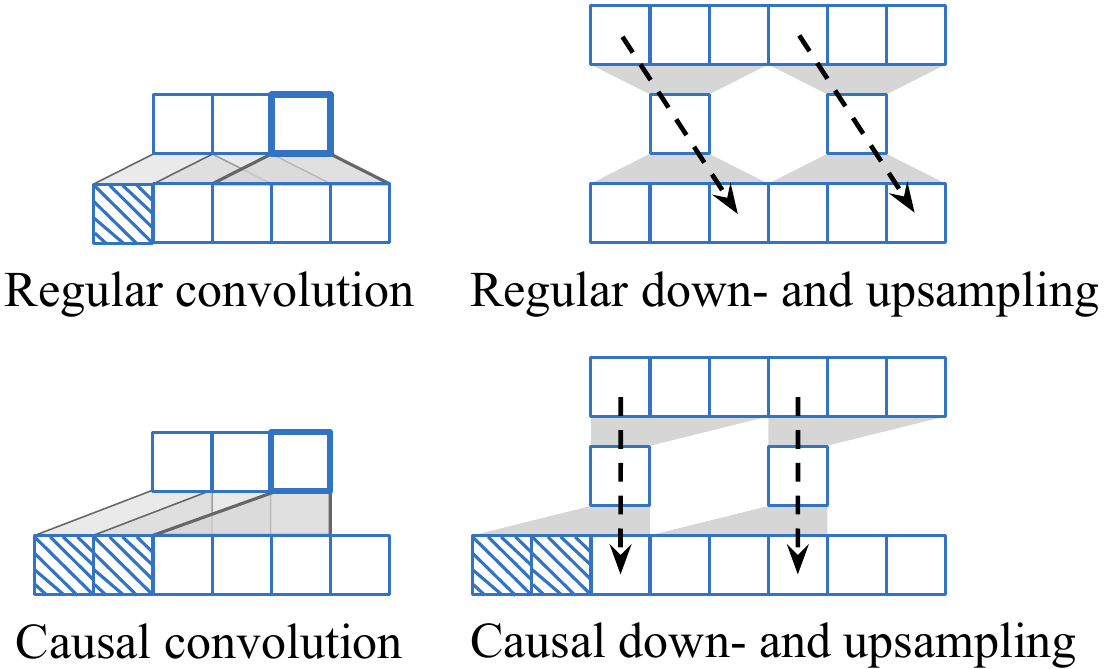}%
                    \end{adjustbox}
                }
                \\
                \captionsetup[subfloat]{justification=raggedright,singlelinecheck=false,margin=-1.0cm}
                \subfloat[Alternate attention masking \label{fig:transformer-offline}]{%
                    \begin{adjustbox}{clip,trim=0.05cm 0.0cm 0.0cm 0.0cm,width=0.37\columnwidth,keepaspectratio}
                        \includegraphics{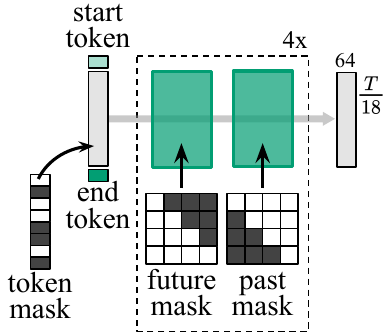}%
                    \end{adjustbox}
                } 
            \end{tabular}
        }
    \end{tabular}
    \caption{(a) TUNeS Transformer block. 
    (b) Illustration of regular and causal convolutions (left) and downsampling operations (right). Padded elements are depicted as cross-hatched squares. Grey trapezoids relate inputs to outputs and thus indicate dependencies ($\dashrightarrow$). Sequences are processed from bottom to top.
    (c) In offline mode, TUNeS Transformer blocks alternately process information from the past or from the future. 
  }
\end{figure}
\endgroup

\subsection{Temporal U-Net with self-attention}
\label{sec:tunes}

For temporal modeling, we propose to combine global self-attention to model long-range dependencies and a temporal convolutional U-Net to introduce local inductive bias.

By design, a U-Net extracts a hierarchy of multi-scale feature maps. We position Transformer blocks at the bottleneck after the U-Net encoder to compute self-attention on the highest-level features, which presumably are pre-filtered and semantically meaningful and thus adequate for attention mechanisms.
Here, the feature maps are also temporally downsampled by an order of magnitude so that the computational cost of attention is acceptable. In addition, global dependencies are analyzed at the core of the U-Net and can be refined further in the upsampling path. This is different in Trans-SVNet and SAHC, which integrate attention towards the end of the model. 
Notably, we do not constrain the flexibility and expressivity of attention by using regularization losses (like OperA, TUT) or local attention only (like Trans-SVNet, ASFormer, TUT). 

The proposed model is a temporal U-Net with self-attention: \emph{TUNeS} (Fig.~\ref{fig:overview}).
Previously, similar architectures were used successfully for image generation \cite{chen2018pixelsnail,ho2020denoising}, image segmentation \cite{petit2021u,rajamani2023attention}, or speech denoising \cite{kong2022speech}.

\subsubsection{Model architecture}
In the encoder, the feature sequence~$\left(f_t\right)_{1 \leq t \leq T}$ is repeatedly passed through temporal convolutions and downsampled. In the decoder, the sequence is repeatedly upsampled and passed through further convolutions. At the bottleneck, \emph{TUNeS Transformer blocks} (Fig.~\ref{fig:transformer-block}) with multi-head self-attention enrich the sequence with global context.
Here, the downsampled feature sequence is treated as a sequence of \emph{tokens}, to which start and end tokens are added.
The TUNeS Transformer block starts with a convolutional block to add \emph{conditional positional encodings} \cite{chu2021conditional}. The \emph{convolutional block} is simply an MS-TCN layer (Fig.~\ref{fig:tcn}) with GELU \cite{hendrycks2016gaussian} activation.

In the decoder, the intermediate feature sequence is enriched with the information of higher-resolution feature maps by means of \emph{skip connections} to the encoder. Here, the sum in each skip connection is weighted with a learnable scalar. 
Further, linear classifier heads after each decoder block compute phase predictions at multiple temporal scales. 

Encoder and decoder consist of convolutional blocks. 
Down- und upsampling by factor~\textit{f} is implemented by 1D~convolutions with kernel~\textit{f} and stride~\textit{f}, where the convolution is transposed for upsampling.
Dilated convolutions ($d=18$) are used in the decoder to retain a large receptive field when progressing to higher-resolution sequences. We set $d=18$ since the coarsest feature maps are downsampled by this factor. 

TUNeS does not include any normalization layers. 
Each operation, including each attention head, processes and outputs sequences with $\mathrm{dim} = 64$ channels.

\subsubsection{Causal operations for online recognition}
To be applicable to online recognition, it is required that the temporal model does not access future information. 
Thus, TUNeS uses causal convolutions for online recognition like TeCNO \cite{czempiel2020tecno}. In addition, causal self-attention is computed in the Transformer blocks by applying a causal mask to the attention weights. 

It is important to perform the downsampling operations in a causal way as well. Like for convolutions, this can be implemented by shifting input sequences by $k - 1$ elements to the right, which corresponds to adding $k - 1$ padded elements at the left (Fig.~\ref{fig:causal-ops}). 
Here, $k$ is the size of the kernel.
In contrast, regular downsampling would leak future information. In that case, an element in the output sequence could depend on up to $k - 1$ future elements in the input sequence.  

\subsubsection{Alternate attention masking for offline recognition}
\label{sec:tunes_attn_masking}
When computing global self-attention in offline mode, it may be difficult to correctly separate past from future information. 
To help with this, we alternately prevent access to future and to past elements in the TUNeS Transformer blocks (Fig.~\ref{fig:transformer-offline}). 
Further, we found it beneficial to increase the number of Transformer blocks for offline recognition from two to eight.

\subsubsection{Training objective}
Let $\left\{ \hat{y}^{(i)} : 1 \leq i \leq 4 \right\}$ be the multi-scale phase predictions, where the length of prediction~$\hat{y}^{(i)}$ is reduced by factor~$\chi_i$ and $\chi = (1, 3, 9, 18)$.
For $2 \leq i \leq 4$, we compute downsampled annotations~$y^{(i)} = (y^{(i)}_s)_{1 \leq s \leq T/\chi_i}$, 
where $y^{(i)}_s \in \{0, 1\}^C$ is a binary vector that summarizes the section $\{y_t : (s - 1)\chi_i < t \leq s \chi_i\}$ of $\chi_i$ frames in the full-resolution annotation: If phase~$p$ appears in the section, the $p$-th entry in $y^{(i)}_s$ is 1 and 0 otherwise. 
Since a section may cover more than one phase, $y^{(i)}_s$ may have more than one entry that is non-zero. Thus, the downsampled annotations are \emph{multi-label} annotations.
 
The training objective $\EuScript{L}_{\text{total}}$ summarizes losses at all scales, see~(\ref{eq:loss}).
We use cross-entropy (CE) loss for the full-resolution prediction~$\hat{y}^{(1)}$. For the lower-resolution predictions, we use binary cross-entropy (BCE) loss to handle the multi-label annotations. 
To counter class imbalance, we use class weights~$\gamma$, which are computed on the training data following the \emph{median frequency balancing} \cite{eigen2015predicting} approach.
In addition, we compute a \emph{smoothing loss} $\EuScript{L}_{\text{smooth}}$ \cite{farha2019ms} on the full-resolution prediction, which is weighted by factor~$\lambda_{\text{smooth}} = 0.15$.
\begin{equation}
    \begin{split}    
    \EuScript{L}_{\text{total}} =\,\, &\EuScript{L}_{\text{phase}}(\hat{y}^{(1)}, y) + \sum_{k=2}^4\EuScript{L}_{\text{BCE}}(\hat{y}^{(k)}, y^{(k)}) \\
    &\EuScript{L}_{\text{phase}}(\hat{y}, y) = \EuScript{L}_{\text{CE}}(\hat{y}, y) + \lambda_{\text{smooth}} \cdot  \EuScript{L}_{\text{smooth}}(\hat{y})
    \end{split}
    \label{eq:loss}
\end{equation}
The individual loss terms are defined in the supplementary.

\subsubsection{Regularization}
\label{sec:sequence_augmentation}
For data augmentation, we augment feature sequences by (i)~shifting the complete sequence to the right by a small random number of time steps and by (ii)~increasing and decreasing the video speed by randomly dropping and duplicating elements of the feature sequence. 

In addition, we mask spans in the token sequences that are input to the Transformer blocks, inspired by the pre-training procedures for masked language models \cite{devlin2018bert,joshi2020spanbert}.
Here, we randomly sample masks to cover spans that correspond to 18--300\,s in the full-resolution video. However, we make sure that phase transitions are not masked because it would be impossible to reconstruct their exact timing.  In total, around 35\% of all tokens are masked per sequence.

\section{Experiments}
\label{sec:experiments}

\subsection{Dataset and evaluation metrics}
Experiments were conducted on the \emph{Cholec80} dataset \cite{twinanda2016endonet}. Cholec80 consists of 80 video recordings of laparoscopic cholecystectomies, which are labeled with phase and tool information. 
The $C=7$ surgical phases are detailed in section~\ref{sec:introduction}.
We used only the phase labels to train feature extractors and temporal models. 
The videos were processed at a temporal resolution of 1\,fps.
Unless stated otherwise, experiments were performed on the 32:8:40 data split, meaning that the first 32 videos in Cholec80 were used for training, the next 8~videos for validation, and the last 40 videos for testing. In particular, hyperparameters were tuned on the validation subset. 

We use the following evaluation metrics:
Video-wise accuracy,
video-wise Macro Jaccard (or Intersection over Union), and
$\mathit{F1}$\,score, which is computed as the harmonic mean of the mean video-wise Precision and the mean video-wise Recall. 
The Macro Jaccard for one test video is the average of the phase-wise Jaccard scores computed on this video. 
To avoid problems with undefined values, we ignore the phase-wise scores for phase~$p$ and video~$v$ if $p$ is not annotated in $v$.

For comparison with prior art, we also report the relaxed evaluation metrics $\EuScript{R}\text{-}\mathsf{Jaccard}$ (or $\EuScript{R}\text{-}\mathsf{Precision}$ and $\EuScript{R}\text{-}\mathsf{Recall}$) and $\EuScript{R}\text{-}\mathsf{Accuracy}$, which were proposed to tolerate certain errors within 10\,s before or after an annotated phase transition.
For all metrics, we used the implementation provided by \cite{funke2023metrics}.

\subsection{Comparisons to baselines and state of the art}
\label{sec:exp_baselines}

We conducted experiments 
\begin{enumerate*}[label=\arabic*)]
  \item to explore the impact of providing longer temporal context when training the visual feature extractor, and
  \item to compare TUNeS to an extensive number of established temporal models.
\end{enumerate*}

\subsubsection{Feature extractors}
\label{sec:exp_features}
We trained feature extractors with context lengths $L$ of 1, 8, or 64 frames (Table~\ref{tab:baselines_online}).
The feature extractor with $L=1$ is simply a ResNet-50, without the LSTM on top (Fig.~\ref{fig:training_no_context}).
Following the custom sampling strategy (section~\ref{sec:temporal_context}), 1264 video sequences were processed in each epoch, for any size of~$L$.
With $L=64$, we trained the feature extractor for 200~epochs. 
For comparability, we trained the feature extractor with $L=8$ for 1600~epochs and with $L=1$ for 12800~epochs. Thus, all feature extractors see the same number of video frames during training, but in different contexts and in different order.
The batch size~$B$ was 10 video sequences with $L=64$, 80 with $L=8$, and 128 video frames with $L = 1$. 

\begin{figure}[t!]    
    \captionsetup[subfloat]{farskip=0cm}
    \vspace{-1.15em}
    \input{tables/funke.t1}

    \captionsetup[subfloat]{farskip=0cm}
    \centering
    \begin{adjustbox}{clip,trim=0.2cm 0.0cm 0.3cm 0.45cm,width=0.95\columnwidth,keepaspectratio}
        \input{figures/funke8.pgf}%
    \end{adjustbox}
    \caption{%
    Visualization of the online recognition results in Table~\ref{tab:baselines_online}.  
    Error bars indicate the standard deviation over repeated experimental runs.
    Please refer to Table~\ref{tab:baselines_online} for the figure legend and numerical results.}%
    \label{fig:comparison_online}
\end{figure}

\begin{figure}[htbp]
    \vspace{-1.15em}
    \captionsetup[subfloat]{farskip=0cm}
    \input{tables/funke.t2}

    \captionsetup[subfloat]{farskip=0cm}
    \centering
    \begin{adjustbox}{clip,trim=0.2cm 0.0cm 0.3cm 0.45cm,width=0.95\columnwidth,keepaspectratio}
        \input{figures/funke9.pgf}%
    \end{adjustbox}
    \caption{%
    Visualization of the offline recognition results in Table~\ref{tab:baselines_offline}.  
    Error bars indicate the standard deviation over repeated experimental runs.
    Please refer to Table~\ref{tab:baselines_offline} for the figure legend and numerical results.}%
    \label{fig:comparison_offline}
\end{figure}

To assess the feature extractor's abilities when $L > 1$, we used the predictions of the jointly trained LSTM: To infer the current phase, we applied the CNN-LSTM to the $L - 1$ previous and to the current frame and took the final prediction. 
In addition, we performed \emph{carry-hidden evaluation (CHE)} \cite{rivoir2022pitfalls}, where the LSTM's hidden state is carried through the video.

To account for the randomness in training deep-learning models, experiments were repeated five times using different random seeds. 

\input{tables/funke.t3}
\input{tables/funke.t4}
\begin{figure}
	\centerline{%
    \begin{adjustbox}{clip,trim=0.1cm 0.08cm 0.13cm 0.14cm,width=0.95\columnwidth,keepaspectratio}
        \input{figures/funke10.pgf}%
    \end{adjustbox}
    }%
	\caption{%
	Performance measurements for processing feature sequences of increasing duration with different temporal models in offline mode.  
	We also present the number of parameters for each temporal model.
	}%
	\label{fig:performance}
\end{figure}

\subsubsection{Temporal models}
\label{sec:exp_temporal_models}

For each context length~$L \in \{1, 8, 64\}$, we trained TUNeS as well as other temporal models on the extracted feature sequences (Tables~\ref{tab:baselines_online} and \ref{tab:baselines_offline}).
The baseline models included an RNN (2-layer GRU), an MS-TCN (TeCNO), and models that integrate attention in different ways.
For comparability, we set the hidden dimension of all models to $\mathrm{dim}=64$. 
To account for randomness, we repeated each experiment five times on each of the five feature extractor instances, yielding 25 experimental runs in total. 

All models were trained for 75 epochs using the \emph{Adam} \cite{kingma2014adam} optimizer and a batch size of one feature sequence. 
Baseline models were trained on the combined loss $\EuScript{L}_{\text{phase}}$
with $\lambda_{\text{smooth}} = 0.15$, see (\ref{eq:loss}). For models with multiple stages, we computed the loss on the output of all stages.
Feature sequence augmentation was applied as described in section~\ref{sec:sequence_augmentation}. Gradients were clipped to have a maximum norm of 1. Finally, we selected the model that achieved the highest Macro Jaccard on the validation set for testing. 

To select an appropriate learning rate~$\eta$ for each model, we ran a quick sweep over $\eta \in \{10^{-4}, 3 \cdot 10^{-4}, 5 \cdot 10^{-4}, 10^{-3}\}$.
Usually, $\eta = 5 \cdot 10^{-4}$ worked well.
If available, we tried to consider the training strategy of the original code base.
Full training details are listed in the supplementary material.

\subsubsection{Performance measurements}
\label{sec:exp_performance}

For TUNeS and for the baselines, we measured model latency and peak GPU memory allocation in offline mode. Here, we used synthetic feature sequences, which are randomly initialized tensors of dimension $T \times 2048$, with a length $T$ of 450, 900, 1800, 3600, and 7200 frames. 
Fig.~\ref{fig:performance} presents the mean of 1000 measurements, after taking 100 measurements for warm-up.

\subsubsection{Comparison to the state of the art}
\label{sec:exp_sota}

We compare the results of TUNeS, trained on features with maximum context $L = 64$, with previously reported results on the 40:40 split (Tables~\ref{tab:sota_online}~and~\ref{tab:sota_offline}).
Here, the feature extractor with $L=64$ and TUNeS were re-trained on all of the first 40 videos in Cholec80. 
Due to the lack of a validation set, we selected the model after the last epoch for testing. To improve convergence, we continuously reduced the learning rate during training, following a \emph{cosine annealing} schedule \cite{loshchilov2016sgdr}. 
For comparisons on the 40:20:20 split, we evaluated the models trained for the 40:40 split on the last 20 videos in Cholec80. 

\subsubsection*{Discussion}

Training the feature extractor in context proved beneficial: The standalone performance of the feature extractor improved as the temporal context increased. In addition,
\textit{almost all temporal models performed better on top of feature extractors that were trained with longer temporal context} (Fig.~\ref{fig:comparison_online} and \ref{fig:comparison_offline}). 
The only exception is TUT, which performed better on features with $L=8$ than on features with $L=64$. 

Apparently, training in context helps the CNN to learn meaningful, contextualized features that hold crucial information for subsequent analysis.
Yet, training in context is simple to implement and does not require additional manual annotations such as tool labels. 
Also, it does not take more time than the standard frame-wise training. The only requirement is sufficient GPU memory (32\,GB in our experiments). 

When evaluating the baselines (Tables~\ref{tab:baselines_online} and \ref{tab:baselines_offline}), we found that the 2-layer GRU and OperA achieved subpar results. 
For the GRU, we suspect that regularization with the smoothing loss and the proposed feature sequence augmentation is not as effective for a recurrent model. 
For OperA, we presume that the model is suffering from a lack of local inductive bias and struggles to filter information from noisy features.  
Yet, OperA improves considerably on contextualized features ($L = 64$), which might be more meaningful and thus easier to analyze with attention. 
Please note that we used a custom implementation of OperA because official code has not been released. 
TUT uses similar building blocks as OperA but performed considerably better by processing multi-scale sequences with local attention, which introduces an inductive bias \cite{du2023dowe}.
Yet, TUT did not reach the performance of ASFormer or TUNeS. 
Since TUT downsamples and upsamples the feature sequence but abandons skip connections in favor of cross-attention, it may lose too much detail information.

\begin{figure*}
	\centerline{%
    \begin{adjustbox}{clip,trim=1.2cm 0.44cm 0.12cm 0.42cm,width=0.9\textwidth,keepaspectratio}
        \input{figures/funke11.pgf}%
    \end{adjustbox}
    }%
	\caption{%
    Impact of integrating Transformer blocks at different positions in the TUNeS model. Left: Results on the Cholec80 dataset (32:8:40 split). Error bars indicate the standard deviation over repeated experimental runs. Right: Performance measurements in offline mode.
    }%
	\label{fig:attn_positioning}
\end{figure*}

\begin{figure}
	\centerline{%
    \begin{adjustbox}{clip,trim=0.18cm 0.45cm 0.23cm 1.12cm,width=0.92\columnwidth,keepaspectratio}
        \input{figures/funke12.pgf}%
    \end{adjustbox}
    }%
	\caption{%
    Impact of attention masking so that Transformer blocks alternately process past and future information in offline mode (Fig.~\ref{fig:transformer-offline}). 
    We present results on the Cholec80 dataset (32:8:40 split). Error bars indicate the standard deviation over repeated experimental runs.
    }%
	\label{fig:attn_masking}
\end{figure}

The best performing baselines were SAHC, TeCNO, TUT, and ASFormer. \textit{TUNeS outperformed the baselines in terms of accuracy and scored better or similarly on Macro Jaccard} (Tables~\ref{tab:baselines_online} and \ref{tab:baselines_offline}).
The results achieved by TUNeS, trained on top of the feature extractor with $L = 64$, also compare favorably to previous results from the literature (Tables~\ref{tab:sota_online}~and~\ref{tab:sota_offline}).
Moreover, \textit{TUNeS is computationally efficient and scales well on long sequences} of up to two hours, regarding both latency and memory consumption (Fig.~\ref{fig:performance}). 

TUNeS is computationally lightweight due to its U-Net structure, where attention is computed only at the bottleneck.
In particular, TUNeS is faster than TeCNO, TUT, and SAHC. 
OperA has even lower latency, but struggles with increased memory usage on long sequences due to the quadratic memory complexity of attention. In contrast, ASFormer uses a custom implementation of local attention to restrict memory consumption, which comes at the cost of increased computation times, especially on long sequences.
Due to the sequential computation, GRU is also relatively slow on long sequences.

\subsection{Ablation studies}
For further insights into TUNeS, additional experiments were conducted on the 32:8:40 split. 
In each ablation, one specific aspect was varied as described below.

\subsubsection{Impact of the position of Transformer blocks}
\label{sec:ablations_positioning}

Originally, Transformer blocks are incorporated at TUNeS block 4, see Fig.~\ref{fig:overview}. 
We studied positioning Transformer blocks at other TUNeS blocks $b$ before ($b \in \{1, 2, 3\}$) 
or after ($b \in \{5, 6, 7\}$) 
the bottleneck. Here, we replaced the convolutional blocks in block~$b$ with $n$ Transformer blocks (online mode: $n=2$, offline mode: $n=8$) and used two convolutional blocks at the bottleneck. Thus, we swapped TUNeS block $b$ and TUNeS block 4, meaning that the overall number of model parameters remained unchanged.
As lower bound, we trained a variant of TUNeS that does not use attention at all. Here, we replaced the $n$~Transformer blocks at the bottleneck with $n$ convolutional blocks. 
For all model configurations, we conducted performance measurements as described in section~\ref{sec:exp_performance}.

\begin{figure*}
	\centerline{%
    \begin{adjustbox}{clip,trim=2cm 0.43cm 2.19cm 0.2cm,width=0.975\textwidth,keepaspectratio}
        \input{figures/funke13.pgf}%
    \end{adjustbox}
    }
	\caption{%
    A 2-stage variant of TUNeS can be used to reduce over-segmentation errors. 
    Left: Offline recognition results on the Cholec80 dataset (32:8:40 split). Error bars indicate the standard deviation over repeated experimental runs. 
    Right: Performance measurements in offline mode.
    }%
	\label{fig:edit_score}
\end{figure*}

\subsubsection*{Discussion}

We hypothesize that Transformer blocks should be positioned at the bottleneck so that attention (i)~is computed on high-level features, and (ii)~is used early so that global context is available throughout the decoder.
The results (Fig.~\ref{fig:attn_positioning}), support our intuition:
If Transformer blocks are positioned further away from the bottleneck, recognition accuracy deteriorates. In offline mode, where more Transformer blocks are used, training becomes more unstable.
Also, attention needs to be computed on higher-resolution sequences, which is more costly regarding latency and GPU memory. 

Notably, using Transformer blocks towards the end of the model, where features are expected to be more meaningful, performs better than using Transformer blocks at the beginning. 
Yet, on highly contextualized features ($L=64$) this effect is less pronounced in online mode. 
Compared to TUNeS, the model with no attention at all performs considerably worse. 
However, using no attention is often better than using attention at ineffective positions, especially in TUNeS blocks 1 and 2. 

\subsubsection{Impact of alternate attention masking} 

We trained and tested models for \emph{offline} recognition either \emph{with} attention masks, which alternately prevent access to future and to past information, or \emph{without} any attention masks. 
Here, we studied three temporal models with global attention: OperA, SAHC, and TUNeS. 
For SAHC, which uses individual Transformer blocks instead of stacks of multiple Transformer blocks (Fig.~\ref{fig:architectures}f), we duplicated each Transformer block to implement alternate attention masking. Here, the first block queries past information and the second block queries future information.

\subsubsection*{Discussion}

Even though attention masks restrict the information that can be accessed in each Transformer block, all temporal models achieved better results \emph{with} attention masks, see Fig.~\ref{fig:attn_masking}.
This applied also to the advanced SAHC model. Here, an additional experiment with duplicated Transformer blocks but without attention masks demonstrates that the improvement does not come from additional model parameters alone.
For OperA, which is lacking local inductive bias, alternate attention masking is actually crucial for acceptable performance. 
Notably, He \emph{et al.} \cite{he2022empirical} also reported difficulties to train a 1-layer Transformer encoder, which is similar to the OperA model, without attention masks 
for offline recognition.

\subsubsection{Quality of predicted phase segments}
\label{sec:ablations_editscore}

The frame-wise annotation $y$ induces a partitioning $Y$ into phase segments,
where each \emph{segment} summarizes all consecutive frames of a phase. 
Likewise, the frame-wise prediction~$\hat{y}$ induces the estimated segmentation $\tilde{Y}$. 
Due to the incorrect classification of individual frames, it is common for $\tilde{Y}$ to divide a video into many fragmented segments. 
To quantify this, Lea \emph{et al.} proposed the \emph{edit score}  \cite{lea2016learning}, which is the normalized Levenshtein distance between $Y$ and $\tilde{Y}$. 
This metric penalizes out-of-order predictions and over-segmentation errors. 

Fig.~\ref{fig:edit_score} shows the mean  video-wise edit score for phase segments predicted in offline mode by TUNeS and other temporal models.
TUNeS obtains higher edit scores than GRU, OperA, and SAHC. Yet, the 2-stage models yield better segmentations, especially TUT. 
We assume that the multi-stage setup -- in addition to the repeated down- and upsampling in TUT -- helps to reduce noise and segment fragmentation. 
Consequently, we also trained a \emph{2-stage variant} of TUNeS, see ({\ref{eq:ms_tunes}}). Here, we optimized the sum of the losses at both stages.
\begin{equation}
    \text{2-stage\,}\mathsf{TUNeS}(f) = \mathsf{TUNeS}\left(\mathrm{Softmax}\left(\mathsf{TUNeS}(f)\right)\right)
    \label{eq:ms_tunes}
\end{equation}
2-stage TUNeS achieves high edit scores that approach those of TUT and, in contrast to TUT, preserves high precision in terms of accuracy and Macro Jaccard. 
However, this improvement comes at the cost of increased computation time (Fig.~\ref{fig:edit_score}, right), which doubles due to the additional TUNeS stage.

In online mode, where the current phase needs to be recognized immediately, with incomplete information, and incorrect decisions cannot be revised, it is difficult to avoid noisy predictions.
Here, initial experiments showed that 2-stage TUNeS could also improve the edit score, but this could come at the cost of decreased accuracy and Macro Jaccard. 
In addition, even the best models with regard to edit score (GRU and 2-stage TUNeS) could not reach a score above $40\,\%$. 

\begin{figure}
	\centerline{%
    \begin{adjustbox}{clip,trim=0.52cm 0.08cm 0.08cm 0.2cm,width=0.95\columnwidth,keepaspectratio}
        \input{figures/funke14.pgf}%
    \end{adjustbox}
    }%
	\caption{%
    Ablation studies on the Cholec80 dataset (32:8:40 split). Unless stated otherwise, we used features with context $L = 64$. Error bars indicate the standard deviation over repeated experimental runs.
    }%
	\label{fig:ablations}
\end{figure}

\begin{figure}[bt]
	\centerline{%
    \begin{adjustbox}{clip,trim=0.07cm 0.07cm 0.23cm 0.1cm,width=0.92\columnwidth,keepaspectratio}
        \input{figures/funke15.pgf}%
    \end{adjustbox}
    }%
	\caption{%
    Impact of the number of Transformer blocks used in TUNeS. We present results on the Cholec80 dataset (32:8:40 split), using features with context $L = 64$. Error bars indicate the standard deviation over repeated experimental runs.
    }%
	\label{fig:nblocks}
\end{figure}

\subsubsection{Additional ablations}
\label{sec:ablations_other}
For $L = 64$, we conducted experiments with the following modifications (Fig.~\ref{fig:ablations} and \ref{fig:nblocks}):

\begin{enumerate*}[label=(\roman*),ref=\roman*]
    \item To explore the importance of conditional positional encodings, we (a) omitted the convolutional block in the TUNeS Transformer block or (b) used standard sinusoidal positional encodings \cite{vaswani2017attention} instead. 
    \item We used regular instead of dilated convolutions in the TUNeS decoder.
    \item \label{abl_regularization} To investigate the effect of measures for regularization, we
    \begin{enumerate*}[label=(\alph*),ref=\alph*]
        \item \label{abl_dropout} trained without dropout, i.e., we deactivated dropout in the convolutional blocks and feedforward networks, 
        \item trained without token masking, or 
        \item trained without feature sequence augmentation.   
    \end{enumerate*}
    \item To further explore the influence of visual features, we
    \begin{enumerate*}[label=(\alph*),ref=\alph*]
        \item extracted the features from the LSTM cell of the CNN-\mbox{LSTM}, using the features right before the linear classifier, or 
        \item used features\footnote{Downloaded from \url{https://github.com/MRUIL/LoViT}.} from a Vision Transformer that was trained with temporal context as described in the \emph{LoViT} paper \cite{liu2023lovit}. 
    \end{enumerate*}
    \item To study the importance of attention, we replaced the TUNeS Transformer blocks with convolutional blocks.
    Additional results for TUNeS without attention are presented in Fig.~\ref{fig:attn_positioning}.
    \item To explore the impact of the number~$n$ of Transformer blocks, we varied $n \in \{ 2, 4, 6, 8, 10\}$. 
\end{enumerate*}

\subsubsection*{Discussion}
As shown in Fig.~\ref{fig:ablations}, most modifications impaired the results, but in many cases the decline was not severe.
\begin{enumerate*}[label=(\roman*),ref=\roman*]
    \item Omitting the convolutional positional encoding did not lead to considerably worse results. Probably, the convolutions in the encoder provide sufficient local inductive bias. Also, sinusoidal positional encodings did not improve results. In online mode, they worked as well as using no positional encodings.
    In offline mode, they caused training instability and overall inferior results. Presumably, this kind of rigid and absolute positional encoding is not suitable for modeling past and future information in sequences of widely varying length. 
    \item Foregoing dilations in the decoder convolutions slightly affected the results for online recognition. The offline model seems to be robust in this regard, possibly due to the acausal operations, which have a symmetric receptive field. 
    \item Regarding different regularization techniques, we found that deactivating dropout had only a minimal effect on online and offline recognition. 
    Dropping token masks worsened the results slightly.
    Omitting feature sequence augmentation had a noticeable negative effect. 
    In particular, Macro Jaccard dropped by 1.8\,\% for online recognition and by  1.1\,\% for offline recognition.  
    \item Using TUNeS with features other than those proposed impaired the results considerably, emphasizing the importance of properly learned visual features.
    \begin{enumerate*}[label=(\alph*)]
        \item When using the LSTM features, some of the detail in the CNN features might be lost. 
        Further, the LSTM features might fit the training data too well, making it difficult for the temporal model to learn generalizable patterns based on that. 
        \item With the LoViT features, TUNeS performed approximately as well as with a CNN that was trained with shorter temporal context, namely $L=8$.
        In this case, the ViT is trained with a context of only 30 frames, which are sampled at equal intervals from the beginning of the current phase up to the current time, thus neglecting phase transitions.   
    \end{enumerate*}
    \item 
    The impact of attention was the greatest: 
    Without attention to model global temporal relationships, 
    Macro Jaccard dropped by 2.4\,\% in online mode and by 4.1\,\% in offline mode. 
    Without attention, TUNeS performed similarly to TeCNO, which might indicate that multi-scale modeling alone cannot explain the performance of TUNeS.
    Rather, \emph{the effective integration of Transformer blocks is a crucial factor}.
    \item When scaling the number of Transformer blocks (Fig.~\ref{fig:nblocks}), we found that using more Transformer blocks was helpful for offline recognition but not for online recognition. Generally, it might be difficult to further improve on the online task due to the inherent uncertainty and ambiguity of this problem.  
    Here, even global information is incomplete because it is only information from the past.
\end{enumerate*}

\begin{figure}[tb]
\centering
    \captionsetup[subfloat]{farskip=0cm}
    \subfloat[Video 47 \label{fig:low}]{        
        \begin{adjustbox}{clip,trim=0.cm 0.cm 0.3cm 0.cm,width=0.9\columnwidth,keepaspectratio}
            \includegraphics{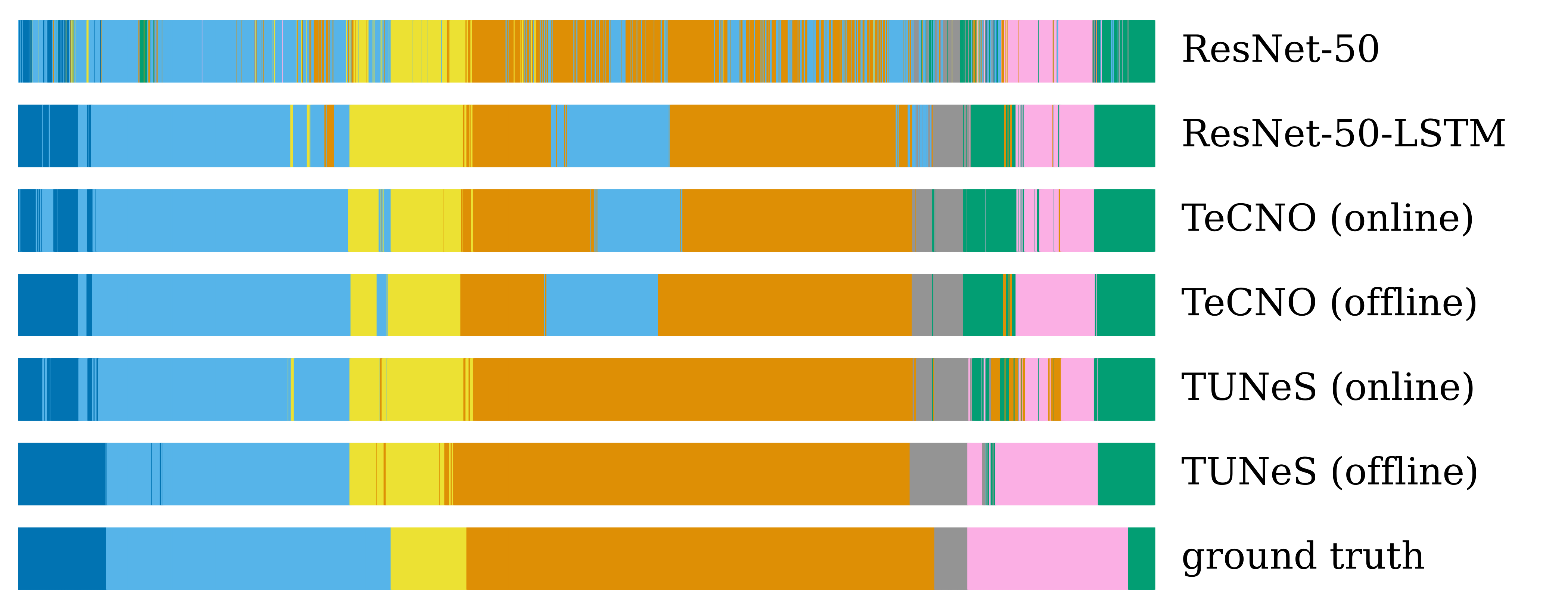}%
        \end{adjustbox}
    }
    
    \subfloat[Video 42 \label{fig:median}]{
        \begin{adjustbox}{clip,trim=0.cm 0.cm 0.3cm 0.cm,width=0.9\columnwidth,keepaspectratio}
            \includegraphics{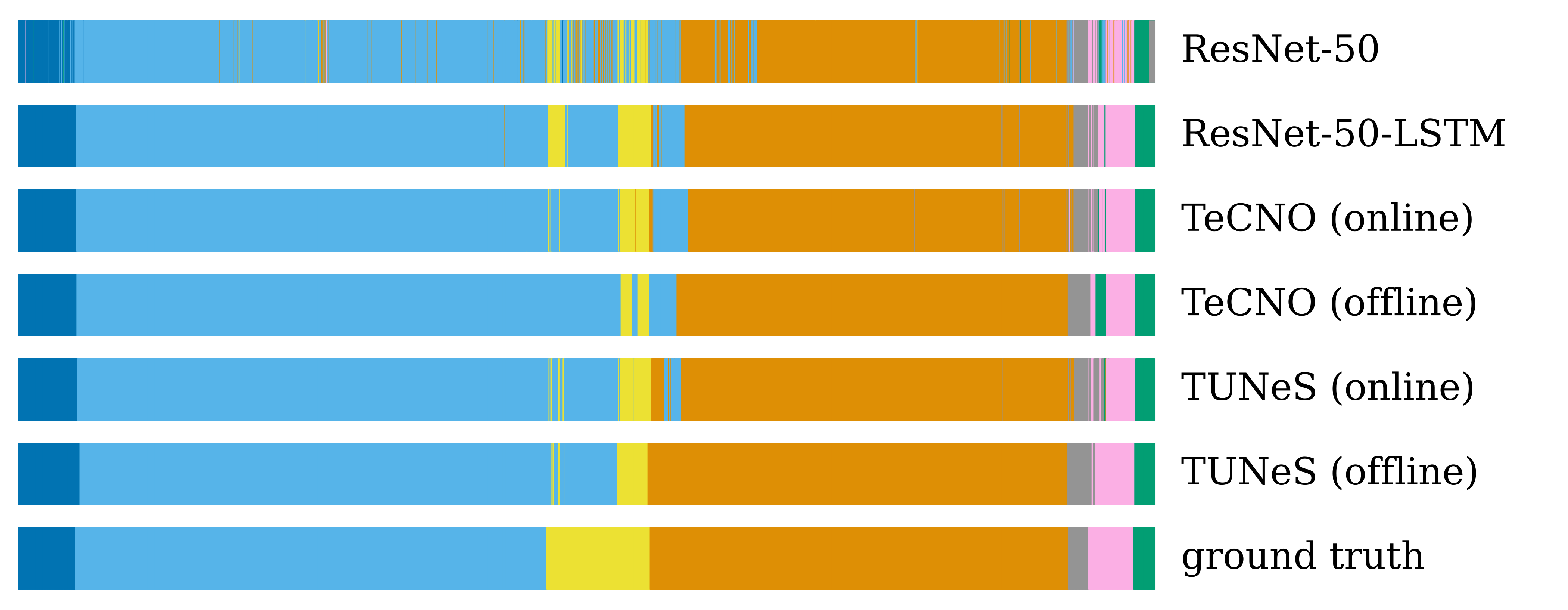}%
        \end{adjustbox}
    }
    
    \subfloat[Video 73 \label{fig:high}]{
        \begin{adjustbox}{clip,trim=0.cm 0.cm 0.3cm 0.cm,width=0.9\columnwidth,keepaspectratio}
            \includegraphics{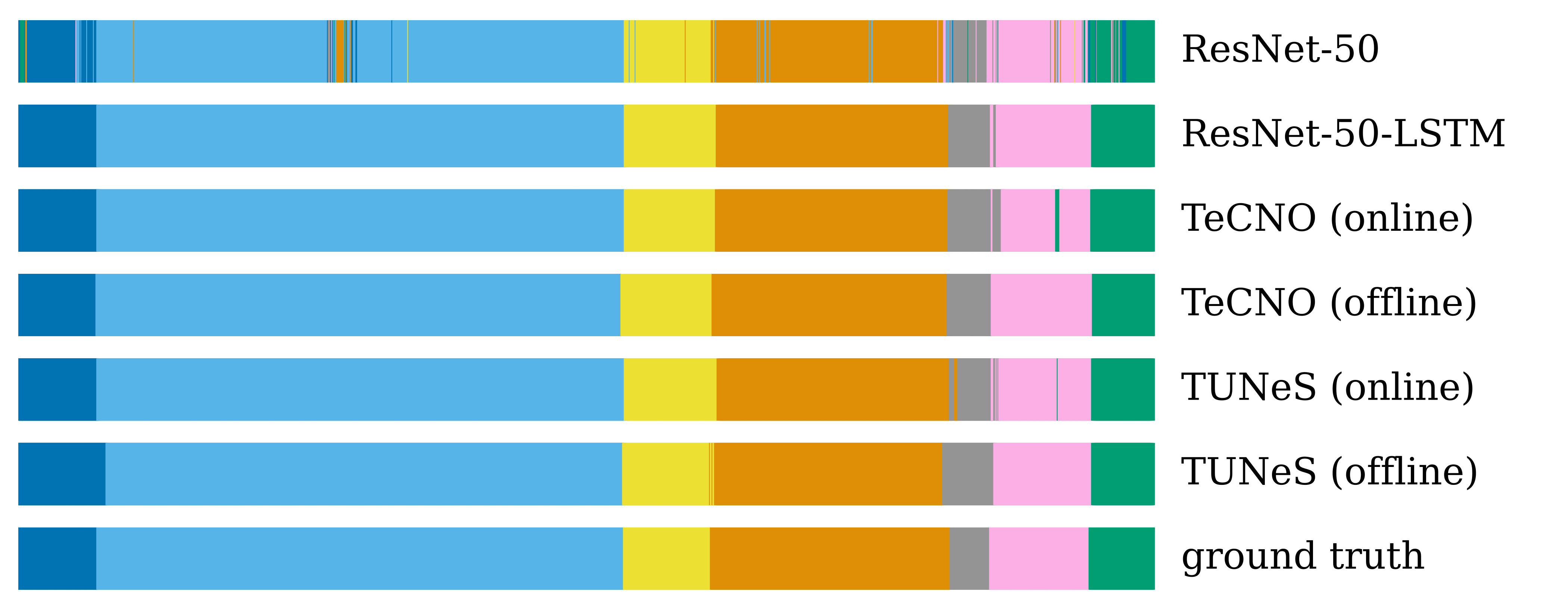}%
        \end{adjustbox}
    }
    \caption{Computed phases on test videos from Cholec80, where the ResNet-50--LSTM achieves the (a) lowest, (b) median, and (c) highest Macro Jaccard. (Surgical phases: \protect\phaseA\,Preparation, \protect\phaseB\,Calot triangle dissection, \protect\phaseC\,Clipping and cutting, \protect\phaseD\,Gallbladder dissection, \protect\phaseE\,Gallbladder packaging, \protect\phaseF\,Cleaning and coagulation, \protect\phaseG\,Gallbladder retraction.)}
    \label{fig:visualization}
\end{figure}

\subsubsection*{Conclusion} 
The ablation studies helped us identify major aspects that are decisive for the performance of TUNeS: 
\begin{itemize}
    \item enriching the temporal U-Net structure with \textit{attention to model long-range relationships},
    \item integrating attention at an appropriate \textit{position} in the U-Net structure, and 
    \item combining TUNeS with a properly trained visual feature extractor.
\end{itemize}
For offline recognition, it is further beneficial 
\begin{itemize}
    \item to use \textit{alternate attention masking} to help separate past and future information and 
    \item to extend TUNeS into a 2-stage model to reduce over-segmentation errors caused by noisy predictions.
\end{itemize}

\begin{figure}[b!]
	\centerline{%
    \begin{adjustbox}{clip,trim=0.05cm 0.05cm 0.13cm 0.25cm,width=0.975\columnwidth,keepaspectratio}
        \input{figures/funke17.pgf}%
    \end{adjustbox}
    }%
	\caption{%
    Results on the AutoLaparo dataset (10:4:7 split).
    We present the performance of the feature extractor (ResNet-50--LSTM) and of TUNeS, trained on features with context $L = 64$. Error bars indicate the standard deviation over repeated experimental runs.
    }%
	\label{fig:autolaparo}
\end{figure}

\input{tables/funke.t5}

\subsection{Qualitative results}

Exemplary phase predictions in online and offline mode are presented in Fig.~\ref{fig:visualization}. 
For comparison, we include predictions of the TeCNO model, which proved to be a solid baseline with similar computational requirements as TUNeS.
TeCNO and TUNeS are trained on ResNet-50 features with temporal context $L=64$. The predictions of the corresponding \mbox{ResNet-50}--LSTM are shown as well. In addition, we include the predictions of a naive frame-wise image classifier (\mbox{ResNet-50}, $L=1$).
To show representative examples, we present results on the videos where the ResNet-50--LSTM achieved the lowest, a median, and the highest Macro Jaccard.

Due to the lack of temporal context, the frame-wise ResNet computes noisy and erroneous predictions.
By including short-term temporal context (64 preceding frames), the ResNet-LSTM clearly improves over that.
Yet, using a temporal model with global context (TeCNO or TUNeS) yields even better results, especially in offline mode.
Compared to TeCNO, TUNeS is less likely to predict \say{Gallbladder retraction} too early or to incorrectly predict \say{Calot triangle dissection} after the \say{Clipping and cutting} phase has been completed. 
This highlights the advanced long-term temporal modeling capabilities of TUNeS.

\subsection{Experiments on additional datasets}
To demonstrate its general applicability, we evaluated TUNeS on two additional datasets.
In each experiment, we trained five feature extractors and five temporal models per feature extractor as described in section \ref{sec:exp_baselines}.
\subsubsection{AutoLaparo \cite{wang2022autolaparo}}
This dataset consists of 21 video recordings of laparoscopic hysterectomies and provides phase annotations at a resolution of 1~fps. Here, $C=7$ surgical phases are defined, namely
\begin{enumerate*}[label=(\arabic*)]
  \item Preparation,
  \item Dividing ligament and peritoneum,
  \item Dividing uterine vessels and ligament,
  \item Transecting the vagina,
  \item Specimen removal,
  \item Suturing, and
  \item Washing.
\end{enumerate*}
As proposed in \cite{wang2022autolaparo}, we used the first 10 videos for training, the next 4 videos for validation, and the last 7 videos for testing.

We followed section~\ref{sec:exp_features} to train feature extractors with context lengths $L \in \{8, 64\}$. Here, we doubled the number of training epochs to account for the smaller number of training videos and thus sequences that would be sampled per epoch.
On the features with $L=64$, we trained TUNeS exactly as described in section~\ref{sec:exp_temporal_models}, but using 100 training epochs. Here, we trained with learning rate scheduler and selected the model after the last epoch for testing. In addition, we re-trained TUNeS on the combined set of training and validation videos.

The results are presented in Fig.~\ref{fig:autolaparo}.
To compare to \cite{wang2022autolaparo}, we also report $f\text{-}\mathsf{Jaccard} = 1/C \sum_{p=1}^C f\text{-}\mathsf{Jaccard}_p$. Here, the true positives, false positives, and false negatives to compute the phase-wise $f\text{-}\mathsf{Jaccard}_p$ are counted over all frames in the test set instead of individually for each video \cite{funke2023metrics}. 

Training the feature extractor with long temporal context yielded major improvements over the baselines. 
TUNeS, trained on top of the contextualized features, achieved additional improvements over the standalone ResNet-LSTM. Here, we obtained a mean video-wise accuracy of 0.851 and $f\text{-}\mathsf{Jaccard}$ of 0.609 for online recognition, beating the recent SKiT.
Yet, we see potential for improvement on the hysterectomy phase recognition task once more training data becomes available.
For a start, TUNeS clearly benefits from including the videos from the validation set during training.

\subsubsection{SAR-RARP50 \cite{psychogyios2023sar}} 
This dataset was created for the SAR-RARP50 challenge at MICCAI 2022. It consists of 50 video recordings of a suturing step during robot-assisted radical prostatectomy. Here, $C=8$ fine-granular \emph{surgical actions} (including a background class) are annotated at a resolution of 10~fps. 
In contrast to high-level phases, actions (or \emph{surgemes}) are well-defined surgical motion units \cite{lalys2014surgical},   
here defined as   
 \begin{enumerate*}[label=(\arabic*)]
    \item Picking-up the needle,
    \item Positioning the needle tip,
    \item Pushing the needle through the tissue,
    \item Pulling the needle out of the tissue,
    \item Tying a knot,
    \item Cutting the suture, and
    \item Returning/dropping the needle.
 \end{enumerate*} 
We used the first 35 videos for training, the next 5 videos for validation, and, as specified for the challenge, the last 10 videos for testing.
The challenge specifies accuracy and F1@10 \cite{lea2017temporal} as evaluation metrics and uses their geometric mean as overall score.
Here, F1@10 mainly penalizes over-segmentation errors. 

Because surgical action recognition requires modeling fine-grained motion, we followed the best challenge submissions and used a video model as feature extractor: \emph{X3D-M} \cite{feichtenhofer2020x3d}, a well-established 3D~CNN for video classification that expands the ResNet architecture to the spatiotemporal domain.
We trained X3D-M on 16-frame clips that were sampled randomly from the training videos. See the supplementary for details.

On the X3D features, we trained TUNeS as well as a strong baseline: ASFormer.
Because the challenge permits acausal methods and emphasizes segmentation quality, we used the offline 2-stage TUNeS model (section \ref{sec:ablations_editscore}). 
To encourage smooth predictions, we trained TUNeS and ASFormer without class weights, increased the weight $\lambda_{\text{smooth}}$ of the smoothing loss to $1.0$, and selected the model that achieved the best F1@10 on the validation set. 
In addition, we smoothed all model outputs with a standard Gaussian filter ($\sigma=5$).
Unfiltered results are included in the supplementary.

Table~\ref{tab:sarrap50} presents the results.
Our 2-stage TUNeS achieved considerable improvements over the standalone X3D and outperformed ASFormer with regards to all metrics. TUNeS is also competitive with the best challenge submissions, ranking second after SummerLab-AI's approach. 
Yet, the challenge results are not fully comparable as the participants used different video models and different strategies to develop and train their models on the available data.

To conclude, TUNeS achieved promising results on two additional datasets from different domains, which demonstrates its versatility and effectiveness across diverse use cases.

\section{Conclusion}
In this paper, we proposed a simple method for training a standard CNN with longer temporal context to extract more meaningful visual features for surgical phase recognition at no added cost (section~\ref{sec:temporal_context}). We showcased the superior performance of feature extractors with extended context on the Cholec80 and AutoLaparo datasets. 

Moreover, we presented TUNeS (section~\ref{sec:tunes}), a computationally efficient temporal model for surgical phase recognition.
TUNeS combines convolutions at multiple temporal resolutions, which are arranged in a U-Net structure, with Transformer blocks to model long-range temporal relationships.
Importantly, the Transformer blocks are integrated at the core of the U-Net, where they can be employed most effectively and efficiently. 
In addition, the convolutions encode local inductive bias and the proposed alternate attention masking supports a notion of temporal order in offline mode.
Therefore, TUNeS does require neither hand-crafted local attention patterns nor custom attention regularization losses -- in contrast to previous temporal models that integrated Transformer blocks. 
These results contribute to a deeper understanding of how to leverage attention mechanisms for surgical phase recognition.  

Furthermore, TUNeS achieves strong results on three distinct and clinically relevant temporal segmentation tasks, namely
phase recognition in laparoscopic cholecystectomy (Cholec80)
and laparoscopic hysterectomy (AutoLaparo), and
action recognition in robot-assisted suturing (SAR-RARP50).
Improving the accuracy of automatic surgical workflow recognition is an important step to further advance computer-assisted surgery and surgical data science.

\bibliographystyle{IEEEtranDOI}
\bibliography{references} 



\section*{Supplementary material}

# Supplementary Material
## TUNeS: A Temporal U-Net with Self-Attention for Video-based Surgical Phase Recognition

Isabel Funke, Dominik Rivoir, Stefanie Krell, and Stefanie Speidel, *Member, IEEE*

### A. Training objective

To train TUNeS, we use the loss function $\mathcal{L}_{\text{total}}$, which is computed on the multi-scale predictions $\hat{y}^{(i)}$:

$$\mathcal{L}_{\text{total}} = \mathcal{L}_{\text{phase}}(\hat{y}^{(1)}, y) + \sum_{k=2}^{4} \mathcal{L}_{\text{BCE}}(\hat{y}^{(k)}, y^{(k)}) \quad (1)$$

$$\mathcal{L}_{\text{phase}}(\hat{y}, y) = \mathcal{L}_{\text{CE}}(\hat{y}, y) + \lambda_{\text{smooth}} \cdot \mathcal{L}_{\text{smooth}}(\hat{y})$$

The individual loss terms are defined as follows:

$$\mathcal{L}_{\text{CE}}(\hat{y}, y) = -\frac{1}{T} \sum_{t=1}^{T} \gamma_{y_t} \cdot \log S(\hat{y}_t)_{y_t} \quad (2)$$

$$\mathcal{L}_{\text{smooth}}(\hat{y}) = \frac{1}{(T-1)C} \sum_{t=2}^{T} \sum_{p=1}^{C} \min(\Delta_{t,p}, \tau)^2, \text{ where} \quad (3)$$
$$\Delta_{t,p} = |\log S(\hat{y}_t)_p - \text{sg}(\log S(\hat{y}_{t-1})_p)| \text{ and } \tau = 4$$

$$\mathcal{L}_{\text{BCE}}(\hat{y}, y) = -\frac{1}{TC} \sum_{t=1}^{T} \sum_{p=1}^{C} (\gamma_p \cdot y_{t,p} \cdot \log(\sigma(\hat{y}_{t,p})) + (1 - y_{t,p}) \cdot \log(1 - \sigma(\hat{y}_{t,p}))) \quad (4)$$

Here, $\sigma$ refers to the sigmoid function, $S$ refers to the Softmax function, i.e., $S(\hat{y}_t)_p = \exp \hat{y}_{t,p} / \sum_{c=1}^{C} \exp \hat{y}_{t,c}$, sg refers to the stopgradient operator, which turns its operand into a constant with zero gradients, and $\gamma_p$, $1 \leq p \leq C$, refers to the class weight for phase $p$.

### B. Model and training details

*1) Feature extractor:* We used the AdamW optimizer *AdamW* optimizer [1] and a *one-cycle lr scheduler* [2] with a maximum learning rate of $3 \cdot 10^{-4}$ and a weight decay of 0.01. The model after the last training epoch was used for feature extraction. To save on GPU memory, we finetuned only the upper two blocks (conv4_x and conv5_x) of the ResNet-50. Video frames were obtained at a resolution of $400 \times 225$ pixels and then cropped to $384 \times 224$ pixels. We used the *Albumentations* library [3] to apply stochastic transformations to the video frames, including color and contrast modifications, blurring, geometric transformations, and horizontal flipping.

*2) TUNeS:* We trained with learning rate $\eta = 5 \cdot 10^{-4}$ ($10^{-3}$ for online recognition on AutoLaparo). All feature sequences were padded to a length that is divisible by 18.

If an experiment involved comparisons to baseline models, the model with highest Macro Jaccard on the validation set was selected for testing. Otherwise, we trained with *cosine annealing* learning rate (*lr*) scheduler and tested the model after the final epoch. This way, the results on the 32:8:40 split improved slightly (see Tables I and II).

In few ablation experiments we observed training instability, which we could reduce by deactivating dropout in all convolutional blocks and feedforward networks.

*3) GRU [4]:* We used a GRU network with two layers and trained with learning rate $\eta = 10^{-3}$. For offline recognition, we used a bidirectional 2-layer GRU.

*4) TeCNO [5]:*[1] We used two stages with 9 layers each in online mode and, to increase the size of the receptive field, 11 layers in offline mode.

*5) ASFormer [6] (offline only):*[2] For comparability with TeCNO, we used one decoder to obtain two stages in total. Per stage, we used 11 layers. To conform with the original codebase, we trained with a weight decay of $10^{-5}$ and applied channel dropout ($p = 0.3$) to the input features.

*6) SAHC [7]:*[3] We used 11 layers in the first stage and 10 layers in the following three stages. As in the original code base, we trained SAHC for 100 epochs with a weight decay of $10^{-5}$ and an initial *lr* of $5 \cdot 10^{-4}$, which was halved every 30 epochs. We used $\lambda_{\text{smooth}} = 1.0$, no class weights in the cross-entropy loss, no gradient clipping, and channel dropout ($p = 0.5$) on the input features. Please note that we adjusted SAHC to use causal downsampling operations and causal attention in online mode and acausal convolutions in offline mode.

*7) OperA [8]:* We implemented OperA as stack of 11 Transformer blocks with one attention head and 256 hidden units in the feedforward network. We trained with $\eta = 10^{-4}$ and the additional attention regularization [8]. We used attention masks in offline mode as proposed for TUNeS.

*8) TUT [9] (offline only):*[4] We used two stages with $2 \cdot 5$ layers per stage. We used Transformer blocks with four heads and a

I. Funke, D. Rivoir, S. Krell, and S. Speidel are with the Department of Translational Surgical Oncology, National Center for Tumor Diseases (NCT/UCC) Dresden, Dresden, Germany; German Cancer Research Center (DKFZ), Heidelberg, Germany; Faculty of Medicine and University Hospital Carl Gustav Carus, TUD Dresden University of Technology, Dresden, Germany; Helmholtz-Zentrum Dresden-Rossendorf (HZDR), Dresden, Germany (e-mail: isabel.funke@nct-dresden.de).

I. Funke, D. Rivoir, and S. Speidel are with the Centre for Tactile Internet with Human-in-the-Loop (CeTI), TUD Dresden University of Technology, Dresden, Germany.

S. Krell and S. Speidel are with the BMBF Research Hub 6G-Life, TUD Dresden University of Technology, Dresden, Germany.

---

[1] Code based on https://github.com/yabufarha/ms-tcn
[2] Code based on https://github.com/ChinaYi/ASFormer
[3] Code based on https://github.com/xmed-lab/SAHC
[4] Code based on https://github.com/ddz16/TUT

TABLE I
ONLINE RECOGNITION ON CHOLEC80 (32:8:40 SPLIT) USING FEATURE EXTRACTORS WITH DIFFERENT TEMPORAL CONTEXT $L$.

| Model | Accuracy[a] | Macro Jaccard[a] |
|---|---|---|
| $L = 1$ | | |
| ResNet-50 | 0.816 ± 0.003 | 0.591 ± 0.005 |
| TeCNO* | 0.877 ± 0.006 | 0.705 ± 0.016 |
| Trans-SVNet* | 0.879 ± 0.007 | 0.705 ± 0.018 |
| GRU | 0.875 ± 0.005 | 0.701 ± 0.011 |
| OperA | 0.863 ± 0.009 | 0.659 ± 0.013 |
| SAHC | 0.882 ± 0.006 | 0.718 ± 0.017 |
| TeCNO | <u>0.891 ± 0.004</u> | <u>0.730 ± 0.009</u> |
| TUNeS | **0.901 ± 0.009** | **0.746 ± 0.013** |
| w/ *lr* scheduler[b] | 0.898 ± 0.009 | 0.743 ± 0.010 |
| $L = 8$ | | |
| ResNet-50-LSTM | 0.865 ± 0.003 | 0.693 ± 0.007 |
| w/ *CHE* | 0.878 ± 0.007 | 0.713 ± 0.012 |
| TeCNO* | 0.883 ± 0.005 | 0.721 ± 0.011 |
| Trans-SVNet* | 0.889 ± 0.008 | 0.723 ± 0.027 |
| GRU | 0.879 ± 0.005 | 0.704 ± 0.011 |
| OperA | 0.880 ± 0.008 | 0.680 ± 0.013 |
| SAHC | 0.892 ± 0.012 | 0.736 ± 0.017 |
| TeCNO | <u>0.900 ± 0.004</u> | <u>0.750 ± 0.008</u> |
| TUNeS | **0.915 ± 0.006** | **0.768 ± 0.010** |
| w/ *lr* scheduler[b] | 0.917 ± 0.005 | 0.773 ± 0.007 |
| $L = 64$ | | |
| ResNet-50-LSTM | 0.901 ± 0.007 | 0.758 ± 0.012 |
| w/ *CHE* | 0.893 ± 0.021 | 0.723 ± 0.026 |
| TeCNO* | 0.898 ± 0.005 | 0.750 ± 0.010 |
| Trans-SVNet* | 0.898 ± 0.008 | 0.749 ± 0.014 |
| GRU | 0.901 ± 0.007 | 0.759 ± 0.012 |
| OperA | 0.898 ± 0.015 | 0.745 ± 0.015 |
| SAHC | <u>0.913 ± 0.010</u> | <u>0.779 ± 0.016</u> |
| TeCNO | 0.906 ± 0.008 | 0.768 ± 0.013 |
| TUNeS | **0.922 ± 0.007** | **0.783 ± 0.011** |
| w/ *lr* scheduler[b] | 0.923 ± 0.004 | 0.784 ± 0.007 |

[a] The reported sample standard deviation refers to the variation over repeated experimental runs.
[b] With *cosine annealing* learning rate scheduler, testing the model after the final training epoch.
*No feature sequence augmentation, no smoothing loss.

TABLE II
OFFLINE RECOGNITION ON CHOLEC80 (32:8:40 SPLIT) USING FEATURE EXTRACTORS WITH DIFFERENT TEMPORAL CONTEXT $L$.

| Model | Accuracy[a] | Macro Jaccard[a] |
|---|---|---|
| $L = 1$ | | |
| GRU | 0.881 ± 0.007 | 0.708 ± 0.011 |
| OperA | 0.873 ± 0.009 | 0.677 ± 0.014 |
| SAHC | 0.898 ± 0.004 | 0.750 ± 0.011 |
| TeCNO | 0.909 ± 0.005 | 0.779 ± 0.012 |
| TUT | 0.913 ± 0.005 | 0.788 ± 0.014 |
| ASFormer | <u>0.918 ± 0.004</u> | <u>0.797 ± 0.011</u> |
| TUNeS | **0.927 ± 0.016** | **0.799 ± 0.026** |
| w/ 2 stages | 0.931 ± 0.010 | 0.803 ± 0.022 |
| w/ *lr* scheduler[b] | 0.929 ± 0.009 | 0.799 ± 0.016 |
| $L = 8$ | | |
| GRU | 0.902 ± 0.005 | 0.749 ± 0.011 |
| OperA | 0.890 ± 0.007 | 0.703 ± 0.010 |
| SAHC | 0.909 ± 0.004 | 0.775 ± 0.008 |
| TeCNO | 0.917 ± 0.003 | 0.798 ± 0.007 |
| TUT | 0.918 ± 0.004 | 0.795 ± 0.007 |
| ASFormer | <u>0.922 ± 0.004</u> | <u>0.805 ± 0.008</u> |
| TUNeS | **0.938 ± 0.007** | **0.816 ± 0.013** |
| w/ 2 stages | 0.941 ± 0.007 | 0.824 ± 0.014 |
| w/ *lr* scheduler[b] | 0.939 ± 0.010 | 0.817 ± 0.016 |
| $L = 64$ | | |
| GRU | 0.917 ± 0.007 | 0.788 ± 0.012 |
| OperA | 0.924 ± 0.008 | 0.784 ± 0.016 |
| SAHC | 0.920 ± 0.006 | 0.804 ± 0.011 |
| TeCNO | 0.920 ± 0.008 | 0.807 ± 0.011 |
| TUT | 0.915 ± 0.006 | 0.791 ± 0.008 |
| ASFormer | <u>0.928 ± 0.005</u> | **0.824 ± 0.008** |
| TUNeS | **0.942 ± 0.006** | <u>0.823 ± 0.015</u> |
| w/ 2 stages | 0.940 ± 0.009 | 0.823 ± 0.016 |
| w/ *lr* scheduler[b] | 0.946 ± 0.004 | 0.825 ± 0.010 |

[a] The reported sample standard deviation refers to the variation over repeated experimental runs.
[b] With *cosine annealing* learning rate scheduler, testing the model after the final training epoch.

local attention window of 31. We trained for 100 epochs with a weight decay of $10^{-5}$ and channel dropout ($p = 0.4$) on the input features. We added the boundary-aware loss [9].

*9) Trans-SVNet [10] (online only):* [5] We trained the Trans-SVNet blocks on top of frozen TeCNO models with 2·9 layers. Following the original code base, we trained Trans-SVNet for 25 epochs with $\eta = 10^{-3}$, using no class weights and no gradient clipping. In each run, the TeCNO model and the Trans-SVNet with the best validation accuracy were selected. Both TeCNO and Trans-SVNet were trained without feature sequence augmentation and without smoothing loss.

*C. Additional results*

Tables I and II provide numerical results for the baseline comparisons, which were presented visually in the main paper.

[5] Code based on https://github.com/xjgaocs/Trans-SVNet



\end{document}

%% file: figure_definitions.tex
\definecolor{darkblue}{rgb}{0.003922, 0.450980, 0.698039}
\definecolor{lightblue}{rgb}{0.337255, 0.705882, 0.913725}
\definecolor{brown}{rgb}{0.792157, 0.568627, 0.380392}
\definecolor{gold}{rgb}{0.870588, 0.560784, 0.019608}
\definecolor{gray}{rgb}{0.580392, 0.580392, 0.580392}
\definecolor{rose}{rgb}{0.984314, 0.686275, 0.894118}
\definecolor{pink}{rgb}{0.800000, 0.470588, 0.737255}
\definecolor{orange}{rgb}{0.835294, 0.368627, 0.000000}
\definecolor{green}{rgb}{0.007843, 0.619608, 0.450980}

\newcommand{\featureExtractor}{\tikz{\draw[line width=1pt, darkblue, fill=darkblue] (-0.8mm,-0.8mm) rectangle (0.8mm,0.8mm);}}
\newcommand{\CHE}{\tikz{\draw[line width=1pt, lightblue, fill=white] (-0.8mm,-0.8mm) rectangle (0.8mm,0.8mm);}}
\newcommand{\TeCNOPlain}{\tikz{\draw[line width=1pt, brown, fill=white] (0,0) circle (0.8mm);}}
\newcommand{\TransSV}{\tikz{\draw[line width=1pt, gold, fill=white] (0,0) circle (0.8mm);}}
\newcommand{\GRU}{\tikz{\draw[line width=1pt, gray] (-0.8mm,0) -- (0.8mm,0) (-0,0.8mm) -- (0,-0.8mm);}}
\newcommand{\OperA}{\tikz{\draw[line width=1pt, rose, fill=white] (0, -0.8mm) -- (0.65mm, 0) -- (0, 0.8mm) -- (-0.65mm, 0) -- cycle;}}
\newcommand{\SAHC}{\tikz{\draw[line width=1pt, pink, fill=pink] (0, -0.8mm) -- (0.65mm, 0) -- (0, 0.8mm) -- (-0.65mm, 0) -- cycle;}}
\newcommand{\TeCNO}{\tikz{\draw[line width=1pt, orange, fill=orange] (0,0) circle (0.8mm);}}
\newcommand{\TUT}{\tikz{\draw[line width=1pt, brown, fill=white] (0, 0.8mm) -- (-0.8mm, -0.8mm) -- (0.8mm, -0.8mm) -- cycle;}}
\newcommand{\ASFormer}{\tikz{\draw[line width=1pt, gold, fill=gold] (0, 0.8mm) -- (-0.8mm, -0.8mm) -- (0.8mm, -0.8mm) -- cycle;}}
\newcommand{\TUNeS}{\tikz{
\draw[line width=1pt, green] (-0.8mm, -0.8mm) -- (0.8mm, 0.8mm); 
\draw[line width=1pt, green] (-0.8mm, 0.8mm) -- (0.8mm, -0.8mm);
}}

\newcommand{\phaseA}{\raisebox{-0.15mm}{\tikz{\draw[line width=1pt, darkblue, fill=darkblue] (-0.3mm,-1mm) rectangle (0.3mm,1mm);}}}
\newcommand{\phaseB}{\raisebox{-0.15mm}{\tikz{\draw[line width=1pt, lightblue, fill=lightblue] (-0.3mm,-1mm) rectangle (0.3mm,1mm);}}}
\newcommand{\phaseC}{\raisebox{-0.15mm}{\tikz{\draw[line width=1pt, yellow, fill=yellow] (-0.3mm,-1mm) rectangle (0.3mm,1mm);}}}
\newcommand{\phaseD}{\raisebox{-0.15mm}{\tikz{\draw[line width=1pt, gold, fill=gold] (-0.3mm,-1mm) rectangle (0.3mm,1mm);}}}
\newcommand{\phaseE}{\raisebox{-0.15mm}{\tikz{\draw[line width=1pt, gray, fill=gray] (-0.3mm,-1mm) rectangle (0.3mm,1mm);}}}
\newcommand{\phaseF}{\raisebox{-0.15mm}{\tikz{\draw[line width=1pt, rose, fill=rose] (-0.3mm,-1mm) rectangle (0.3mm,1mm);}}}
\newcommand{\phaseG}{\raisebox{-0.15mm}{\tikz{\draw[line width=1pt, green, fill=green] (-0.3mm,-1mm) rectangle (0.3mm,1mm);}}}

%% file: tables/funke.t1.tex
\begin{table}[H]
	\caption{Online recognition on Cholec80 (32:8:40 split) using \\ feature extractors with different temporal context~$L$. \\ We present the feature extractor's standalone performance (\protect\featureExtractor, \protect\CHE) as well as the performance of various temporal models, trained on top of the frozen feature extractor. The asterisk* indicates training without feature sequence augmentation and smoothing loss.}
	\label{tab:baselines_online}
	\centerline{
	\begin{tabular}{c@{\hspace{4pt}}lll}
		\toprule
		\multicolumn{2}{l}{Model} & Accuracy$^{\mathrm{a}}$ & Macro Jaccard$^{\mathrm{a}}$ \\ 
		\midrule
		\multicolumn{2}{l}{$L = 1$} & & \\ 
		\cmidrule{1-2} 
		\featureExtractor & ResNet-50 & 0.816 $\pm$ 0.003 & 0.591 $\pm$ 0.005 \\ 
		\TeCNOPlain & TeCNO$^*$ & 0.877 $\pm$ 0.006 & 0.705 $\pm$ 0.016 \\ 
		\TransSV & Trans-SVNet$^*$ & 0.879 $\pm$ 0.007 & 0.705 $\pm$ 0.018 \\ 
		\GRU & GRU & 0.875 $\pm$ 0.005 & 0.701 $\pm$ 0.011 \\ 
		\OperA & OperA & 0.863 $\pm$ 0.009 & 0.659 $\pm$ 0.013 \\ 
		\SAHC & SAHC & 0.882 $\pm$ 0.006 & 0.718 $\pm$ 0.017 \\ 
		\TeCNO & TeCNO & \underline{0.891} $\pm$ \underline{0.004} & \underline{0.730} $\pm$ \underline{0.009} \\ 
		\TUNeS & TUNeS & \textbf{0.901} $\pm$ \textbf{0.009} & \textbf{0.746} $\pm$ \textbf{0.013} \\ 
		\midrule
		\multicolumn{2}{l}{$L = 8$} & & \\ 
		\cmidrule{1-2} 
		\featureExtractor & ResNet-50--LSTM & 0.865 $\pm$ 0.003 & 0.693 $\pm$ 0.007 \\ 
		\CHE & ~~w/ \emph{CHE} & 0.878 $\pm$ 0.007 & 0.713 $\pm$ 0.012 \\ 
		\TeCNOPlain & TeCNO$^*$ & 0.883 $\pm$ 0.005 & 0.721 $\pm$ 0.011 \\ 
		\TransSV & Trans-SVNet$^*$ & 0.889 $\pm$ 0.008 & 0.723 $\pm$ 0.027 \\ 
		\GRU & GRU & 0.879 $\pm$ 0.005 & 0.704 $\pm$ 0.011 \\ 
		\OperA & OperA & 0.880 $\pm$ 0.008 & 0.680 $\pm$ 0.013 \\ 
		\SAHC & SAHC & 0.892 $\pm$ 0.012 & 0.736 $\pm$ 0.017 \\ 
		\TeCNO & TeCNO & \underline{0.900} $\pm$ \underline{0.004} & \underline{0.750} $\pm$ \underline{0.008} \\ 
		\TUNeS & TUNeS & \textbf{0.915} $\pm$ \textbf{0.006} & \textbf{0.768} $\pm$ \textbf{0.010} \\ 
		\midrule
		\multicolumn{2}{l}{$L = 64$} & & \\ 
		\cmidrule{1-2} 
		\featureExtractor & ResNet-50--LSTM & 0.901 $\pm$ 0.007 & 0.758 $\pm$ 0.012 \\ 
		\CHE & ~~w/ \emph{CHE} & 0.893 $\pm$ 0.021 & 0.723 $\pm$ 0.026 \\ 
		\TeCNOPlain & TeCNO$^*$ & 0.898 $\pm$ 0.005 & 0.750 $\pm$ 0.010 \\ 
		\TransSV & Trans-SVNet$^*$ & 0.898 $\pm$ 0.008 & 0.749 $\pm$ 0.014 \\ 
		\GRU & GRU & 0.901 $\pm$ 0.007 & 0.759 $\pm$ 0.012 \\ 
		\OperA & OperA & 0.898 $\pm$ 0.015 & 0.745 $\pm$ 0.015 \\ 
		\SAHC & SAHC & \underline{0.913} $\pm$ \underline{0.010} & \underline{0.779} $\pm$ \underline{0.016} \\ 
		\TeCNO & TeCNO & 0.906 $\pm$ 0.008 & 0.768 $\pm$ 0.013 \\ 
		\TUNeS & TUNeS & \textbf{0.922} $\pm$ \textbf{0.007} & \textbf{0.783} $\pm$ \textbf{0.011} \\ 
		\bottomrule
		\multicolumn{4}{l}{\makecell[cl]{$^{\mathrm{a}}$The reported sample standard deviation refers to \\ the variation over repeated experimental runs.}} 
	\end{tabular}
	}
\end{table}

%% file: tables/funke.t2.tex
\begin{table}[H]
	\caption{Offline recognition on Cholec80 (32:8:40 split) using feature extractors with different temporal context~$L$. \\ We present the performance of various temporal models, trained on top of the frozen feature extractor.}
	\label{tab:baselines_offline}
	\centerline{
	\begin{tabular}{c@{\hspace{4pt}}lll}
		\toprule
		\multicolumn{2}{l}{Model} & Accuracy$^{\mathrm{a}}$ & Macro Jaccard$^{\mathrm{a}}$ \\ 
		\midrule
		\multicolumn{2}{l}{$L = 1$} & & \\ 
		\cmidrule{1-2} 
		\GRU & GRU & 0.881 $\pm$ 0.007 & 0.708 $\pm$ 0.011 \\ 
		\OperA & OperA & 0.873 $\pm$ 0.009 & 0.677 $\pm$ 0.014 \\ 
		\SAHC & SAHC & 0.898 $\pm$ 0.004 & 0.750 $\pm$ 0.011 \\ 
		\TeCNO & TeCNO & 0.909 $\pm$ 0.005 & 0.779 $\pm$ 0.012 \\ 
		\TUT & TUT & 0.913 $\pm$ 0.005 & 0.788 $\pm$ 0.014 \\ 
		\ASFormer & ASFormer & \underline{0.918} $\pm$ \underline{0.004} & \underline{0.797} $\pm$ \underline{0.011} \\ 
		\TUNeS & TUNeS & \textbf{0.927} $\pm$ \textbf{0.016} & \textbf{0.799} $\pm$ \textbf{0.026} \\ 
		\midrule
		\multicolumn{2}{l}{$L = 8$} & & \\ 
		\cmidrule{1-2} 
		\GRU & GRU & 0.902 $\pm$ 0.005 & 0.749 $\pm$ 0.011 \\ 
		\OperA & OperA & 0.890 $\pm$ 0.007 & 0.703 $\pm$ 0.010 \\ 
		\SAHC & SAHC & 0.909 $\pm$ 0.004 & 0.775 $\pm$ 0.008 \\ 
		\TeCNO & TeCNO & 0.917 $\pm$ 0.003 & 0.798 $\pm$ 0.007 \\ 
		\TUT & TUT & 0.918 $\pm$ 0.004 & 0.795 $\pm$ 0.007 \\ 
		\ASFormer & ASFormer & \underline{0.922} $\pm$ \underline{0.004} & \underline{0.805} $\pm$ \underline{0.008} \\ 
		\TUNeS & TUNeS & \textbf{0.938} $\pm$ \textbf{0.007} & \textbf{0.816} $\pm$ \textbf{0.013} \\ 
		\midrule
		\multicolumn{2}{l}{$L = 64$} & & \\ 
		\cmidrule{1-2} 
		\GRU & GRU & 0.917 $\pm$ 0.007 & 0.788 $\pm$ 0.012 \\ 
		\OperA & OperA & 0.924 $\pm$ 0.008 & 0.784 $\pm$ 0.016 \\ 
		\SAHC & SAHC & 0.920 $\pm$ 0.006 & 0.804 $\pm$ 0.011 \\ 
		\TeCNO & TeCNO & 0.920 $\pm$ 0.008 & 0.807 $\pm$ 0.011 \\ 
		\TUT & TUT & 0.915 $\pm$ 0.006 & 0.791 $\pm$ 0.008 \\ 
		\ASFormer & ASFormer & \underline{0.928} $\pm$ \underline{0.005} & \textbf{0.824} $\pm$ \textbf{0.008} \\ 
		\TUNeS & TUNeS & \textbf{0.942} $\pm$ \textbf{0.006} & \underline{0.823} $\pm$ \underline{0.015} \\ 
		\bottomrule
		\multicolumn{4}{l}{\makecell[cl]{$^{\mathrm{a}}$The reported sample standard deviation refers to \\ the variation over repeated experimental runs.}} 
	\end{tabular}
	}
\end{table}

%% file: tables/funke.t3.tex
\begin{table}[htb]
    \captionsetup[subfloat]{farskip=0cm}
	\caption{%
    Online recognition on Cholec80 (40:40 split). Comparison to results from the literature.
    }%
	\label{tab:sota_online}
	\setlength{\tabcolsep}{4pt}
    \centering
    \subfloat{%
	\begin{tabular}{lllll}
		\toprule
		\multirow{2}{*}{Model} & \multicolumn{2}{l}{$\EuScript{R}\text{-}\mathsf{Accuracy}$} & \multicolumn{2}{l}{$\EuScript{R}\text{-}\mathsf{Jaccard}$} \\
            & $M$ & $SD_V$ & $M_P$ & $SD_P$ \\
        \midrule
        \makecell[cl]{ResNet-50--LSTM$^\mathrm{b}$, $L=8$\\ ~~(similar to SV-RCNet\cite{jin2017sv})} & 0.881 & 0.063 & 0.739 & 0.090 \\
        TeCNO \cite{czempiel2020tecno}, \emph{results of} \cite{gao2021trans} & 0.886 & 0.078 & 0.751 & 0.069 \\
        TMRNet w/ ResNeST \cite{jin2021temporal} & 0.901 & 0.076 & 0.791 & 0.057 \\
        Trans-SVNet \cite{gao2021trans} & 0.903 & 0.071 & 0.793 & 0.066 \\
        Not E2E \cite{yi2022not} & 0.920 & 0.053 & 0.771 & 0.115 \\
        SAHC \cite{ding2022exploring} & 0.918 & 0.081 & 0.812 & 0.055 \\
        SKiT w/ ViT \cite{liu2023skit} & 0.934 & 0.053 & 0.826 & -- \\
        ConvNeXt--LSTM$^\mathrm{b}$, $L=256$ \cite{rivoir2022pitfalls} & \underline{0.935} & \underline{0.065} & \underline{0.829} & \underline{0.101} \\
        TUNeS, $L=64$ & \textbf{0.939} & \textbf{0.050} & \textbf{0.842} & \textbf{0.083} \\
		\bottomrule
        \multicolumn{5}{l}{\makecell[cl]{Like prior art, we report the mean~$M$ and the standard deviation over \\ videos~$SD_V$ for $\EuScript{R}\text{-}\mathsf{Accuracy}$. For $\EuScript{R}\text{-}\mathsf{Jaccard}$, we report the mean over \\ phase-wise means~$M_P$ and the standard deviation over phases~$SD_P$.}}
	\end{tabular}
    }%
    \hfill
    \captionsetup[subfloat]{farskip=0.3cm}
    \subfloat{%
	\begin{tabular}{llll}
		\toprule
		Model & Accuracy$^{\mathrm{a}}$ & Macro Jaccard$^{\mathrm{a}}$ & $\mathit{F1}$ score$^{\mathrm{a}}$ \\ 
        \midrule
        \makecell[cl]{ResNet-50--LSTM$^\mathrm{b}$,\\ ~~$L=8$} & 0.872 $\pm$ 0.004 & 0.707 $\pm$ 0.007 & 0.832 $\pm$ 0.003 \\
        TeCNO \cite{czempiel2020tecno} &    &    &    \\
         ~~\emph{results of} \cite{czempiel2022surgical} & 0.874 $\pm$ 0.014 & -- & 0.825 $\pm$ 0.018 \\
         ~~\emph{results of} \cite{chen2022spatio} & 0.900 & 0.723 & 0.839 \\
        Trans-SVNet \cite{gao2021trans} &    &    &     \\
         ~~\emph{results of} \cite{chen2022spatio} & 0.896 & 0.731 & 0.845 \\
        \makecell[cl]{GRU w/\\ ~~VideoSwin \cite{he2022empirical}} & 0.909 & -- & 0.853 \\ 
        \makecell[cl]{PATG w/ SE-\\ ~~ResNet-50 \cite{kadkhodamohammadi2022patg}} & \underline{0.914} & -- & 0.854 \\
        \makecell[cl]{Dual Pyramid\\ ~~ASFormer w/ \\ ~~PCPVT \cite{chen2022spatio}} & \underline{0.914} & \underline{0.754} & \underline{0.858}  \\  
        TUNeS, $L=64$ & \textbf{0.927} $\pm$ \textbf{0.004} & \textbf{0.799} $\pm$ \textbf{0.005} & \textbf{0.890} $\pm$ \textbf{0.003} \\
		\bottomrule
		\multicolumn{4}{l}{\makecell[cl]{$^{\mathrm{a}}$The reported sample standard deviation refers to the variation over \\ repeated experimental runs.}} \\
        \multicolumn{4}{l}{$^{\mathrm{b}}$Trained end-to-end on short video sequences of $L$ frames.} \\ 
	\end{tabular}
    }%
\end{table}

%% file: tables/funke.t4.tex
\begin{table}[htb]
    \captionsetup[subfloat]{farskip=0cm}
	\caption{%
    Offline recognition on Cholec80. Comparison to results from the literature. 
    }%
	\label{tab:sota_offline}
	\setlength{\tabcolsep}{4pt}
    \centering
    \subfloat{%
    \begin{adjustbox}{width=1.02\columnwidth,keepaspectratio}
    \begin{tabular}{llllllll}
		\toprule
		\multirow{2}{*}{Model} & \multicolumn{2}{l}{$\EuScript{R}\text{-}\mathsf{Accuracy}$} & \multicolumn{2}{l}{$\EuScript{R}\text{-}\mathsf{Precision}$} & \multicolumn{2}{l}{$\EuScript{R}\text{-}\mathsf{Recall}$} & split \\
            & $M$ & $SD_V$ & $M_P$ & $SD_P$ & $M_P$ & $SD_P$ & \\
        \midrule
        \makecell[cl]{ASTCFormer \\ ~~\cite{zhang2022surgical}} & 0.957 & 0.033 & 0.923 & 0.062 & 0.912 & 0.095 & 40:20:20 \\ 
        \makecell[cl]{TUNeS,\\ ~~$L=64$} & \textbf{0.964} & \textbf{0.022} & \textbf{0.936} & \textbf{0.070} & \textbf{0.939} & \textbf{0.054} & 40:20:20 \\
		\bottomrule
        \multicolumn{8}{l}{\makecell[cl]{Like prior art, we report the mean~$M$ and the standard deviation over videos \\ $SD_V$ for $\EuScript{R}\text{-}\mathsf{Accuracy}$. For $\EuScript{R}\text{-}\mathsf{Precision}$ and $\EuScript{R}\text{-}\mathsf{Recall}$, we report the mean \\ over phase-wise means~$M_P$ and the standard deviation over phases~$SD_P$.}}
	\end{tabular}
    \end{adjustbox}
    }%
    \hfill
    \captionsetup[subfloat]{farskip=0.3cm}
    \setlength{\tabcolsep}{4pt}
    \subfloat{%
    \begin{adjustbox}{width=1.02\columnwidth,keepaspectratio}
    	\begin{tabular}{lllll}
		\toprule
		Model & Accuracy$^{\mathrm{a}}$ & Macro Jaccard$^{\mathrm{a}}$ & $\mathit{F1}$ score$^{\mathrm{a}}$ & split \\ 
        \midrule
        \makecell[cl]{GRU w/\\ ~~VideoSwin \\ ~~\cite{he2022empirical}} & 0.939 & -- & 0.898 & 40:40 \\ 
        TUNeS, $L=64$ & \textbf{0.949} $\pm$ \textbf{0.005} & 0.836 $\pm$ 0.009 & \textbf{0.912} $\pm$ \textbf{0.005} & 40:40 \\ 
        \makecell[cl]{Transition \\ ~~Retrieval \\ ~~Network \cite{zhang2022retrieval}} & 0.901 & -- & 0.848 & 40:20:20 \\ 
        TUNeS, $L=64$ & \textbf{0.953} $\pm$ \textbf{0.004} & 0.837 $\pm$ 0.011 & \textbf{0.911} $\pm$ \textbf{0.007} & 40:20:20 \\
		\bottomrule
		\multicolumn{5}{l}{\makecell[cl]{$^{\mathrm{a}}$The reported sample standard deviation refers to the variation over repeated \\ experimental runs.}}
	\end{tabular}
    \end{adjustbox}
    }%
\end{table}

%% file: tables/funke.t5.tex
\begin{table*}[tb]
    \caption{Results on the SAR-RARP50 challenge, including the four best submissions (top).\\ We evaluated an X3D video model as feature extractor, combined with ASFormer or 2S-TUNeS as temporal model (bottom). }
	\label{tab:sarrap50}
    \centering
    \begin{tabular}{>{\columncolor{gray!10}}l >{\columncolor{gray!10}}l >{\columncolor{gray!10}}l llll}
    \toprule
    Team & Feature extractor & Temporal model & Accuracy & F1@10 & \makecell[cl]{Geometric \\mean} & Macro Jaccard \\
    \midrule
    NCC-Next \cite{psychogyios2023sar} & VideoSwin + SlowFast & Multi-scale filtering & 0.713 & 0.799 & 0.755 & \\
    CAMI-SIAT \cite{psychogyios2023sar} & Bridge-prompt & ASFormer & 0.770 & 0.806 & 0.788 & \\
    Uniandes \cite{psychogyios2023sar} & MViT & Window-based filtering & 0.786 & 0.823 & 0.804 & \\
    SummerLab-AI \cite{psychogyios2023sar} & Bridge-prompt & ASFormer & \textbf{0.815} & \textbf{0.841} & \textbf{0.828} & \\
     \midrule
                & X3D-M & Gaussian filter & 0.782 $\pm$ 0.008 & 0.731 $\pm$ 0.014 & 0.756 & 0.532 $\pm$ 0.016 \\
                & X3D-M & ASFormer & 0.792 $\pm$ 0.005 &  0.801  $\pm$ 0.008 & 0.796 & 0.553 $\pm$ 0.009 \\
                & X3D-M & 2-stage TUNeS & \underline{0.807} $\pm$ 0.005 & \underline{0.835} $\pm$ 0.015 & \underline{0.821} & 0.607 $\pm$ 0.019 \\
    \bottomrule
    \multicolumn{7}{r}{\cellcolor{white}  \makecell[cr]{The reported sample standard deviation refers to the variation over repeated experimental runs.}} 
    \end{tabular}
\end{table*}